\pdfoutput=1

\documentclass[journal]{IEEEtran}
\usepackage[pscoord]{eso-pic}
\newcommand{\placetextbox}[3]{
\setbox0=\hbox{#3}
\AddToShipoutPictureFG{ \put(\LenToUnit{#1\paperwidth},\LenToUnit{#2\paperheight}){\vtop{{\null}\makebox[0pt][c]{#3}}}}
}

\placetextbox{.47}{0.045}{~\copyright IEEE All rights reserved. Personal use is permitted. For any other purposes permission must be obtained from the IEEE.}

\makeatletter
\let\old@ps@headings\ps@headings
\let\old@ps@IEEEtitlepagestyle\ps@IEEEtitlepagestyle
\def\confheader#1{%
  \def\ps@headings{%
    \old@ps@headings%
    \def\@oddhead{\strut\hfill#1\hfill\strut}%
    \def\@evenhead{\strut\hfill#1\hfill\strut}%
  }%
  \def\ps@IEEEtitlepagestyle{%
    \old@ps@IEEEtitlepagestyle%
    \def\@oddhead{\strut\hfill#1\hfill\strut}%
    \def\@evenhead{\strut\hfill#1\hfill\strut}%
  }%
  \ps@headings%
}
\makeatother

\confheader{%
  This article has been accepted for publication in IEEE/ASME Transactions on Mechatronics.
  DOI 10.1109/TMECH.2021.3081426
}

	\usepackage[pdftex]{graphicx}
	\graphicspath{{Qolo_Docking_2020/Figures/}}
    \DeclareGraphicsExtensions{.pdf} 

  \usepackage{subfigure}
  \usepackage{wrapfig}
  
\usepackage{amsmath}
\DeclareMathOperator{\atantwo}{atan2}
\usepackage{amssymb}
\usepackage{graphicx}
\usepackage{nccmath}
\usepackage{array}

\begin{document}

\title{Virtual Landmark-Based Control of Docking Support for Assistive Mobility Devices}

\author{Yang Chen$^{1 \dagger}$, {Diego} {Paez-Granados}$^{2}$ ~\IEEEmembership{Member,~IEEE,}  Bruno~Leme$^{3}$ ~\IEEEmembership{Member,~IEEE}, and  Kenji~Suzuki$^{4}$~\IEEEmembership{Member,~IEEE}
\thanks{$^\dagger$  is the corresponding author.} 
\thanks{$^{1}$ Y. Chen is with the School of Integrative and Global Majors (SIGMA), University of Tsukuba, Japan.
        {\tt\small chenyang@ai.iit.tsukuba.ac.jp}}  
\thanks{$^{2}${ D.} {Paez-Granados} is with the Ecole Polytechnique Federale de Lausane (EPFL), Switzerland.
        {\tt\small dfpg@ieee.org}}  
\thanks{$^{3}$ B. Leme is with the Faculty of Engineering, Information and Systems, University of Tsukuba, Japan. 
        {\tt\small leme@ieee.org}}  
\thanks{$^{4}$ K. Suzuki is with the Faculty of Engineering and Center for Cybernics Research, University of Tsukuba, Japan. 
        {\tt\small kenji@ieee.org}  
 	}
\thanks{Manuscript received December 31, 2020.}
}
\markboth{IEEE/ASME Transactions on Mechatronics,~Vol.~, No.~, Month~Year}%
{Chen Y. \MakeLowercase{\textit{et al.}}: Autonomous Docking}

\maketitle


\begin{abstract} 			

This work proposes an autonomous docking control for nonholonomic constrained mobile robots and applies it to an intelligent mobility device or wheelchair for assisting the user in approaching resting furniture such as a chair or a bed. We defined a virtual landmark inferred from the target docking destination. Then we solve the problem of keeping the targeted volume inside the field of view (FOV) of a tracking camera and docking to the virtual landmark through a novel definition that enables to control for the desired end-pose. We proposed a nonlinear feedback controller to perform the docking with the depth camera's FOV as a constraint. Then a numerical method is proposed to find the feasible space of initial states where convergence could be guaranteed. 
Finally, the entire system was embedded for real-time operation on a standing wheelchair with the virtual landmark estimation by 3D object tracking with an RGB-D camera and we validated the effectiveness in simulation and experimental evaluations. The results show the guaranteed convergence for the feasible space depending on the virtual landmark location. In the implementation, the robot converges to the virtual landmark while respecting the FOV constraints.

\end{abstract}

\begin{IEEEkeywords}
Human-assistive robot, mobile robot control, autonomous docking, virtual landmark, FOV constraint
\end{IEEEkeywords}

\IEEEpeerreviewmaketitle


\section{INTRODUCTION}\label{sec_intro}
\IEEEPARstart{P}{owered} wheelchairs operation requires docking as one of the daily activities, many times to very specific poses like driving back to face a chair or laterally to a bed so that, the user can transfer between the wheelchair and other surfaces. 
Another type of powered wheelchairs  - standing mobility devices - has been proposed to enable standing mobility for lower-limb impaired users. i.e., the user locomotes in a standing posture rather than seated. One such example is the Qolo device developed by our group \cite{Eguchi2018,Paez2018}. As with powered wheelchairs, controlling the docking manoeuvre to a conventional chair or a bed is usually difficult for individuals with lower-limb disabilities because twisting their upper body while in the device is rather limited or some users might have limited muscle control. Therefore, we believe it is possible to automate this process for assisting end-users that might find difficulties in this task.

\begin{figure}[http]
\begin{center}
\includegraphics[width=0.9\linewidth]{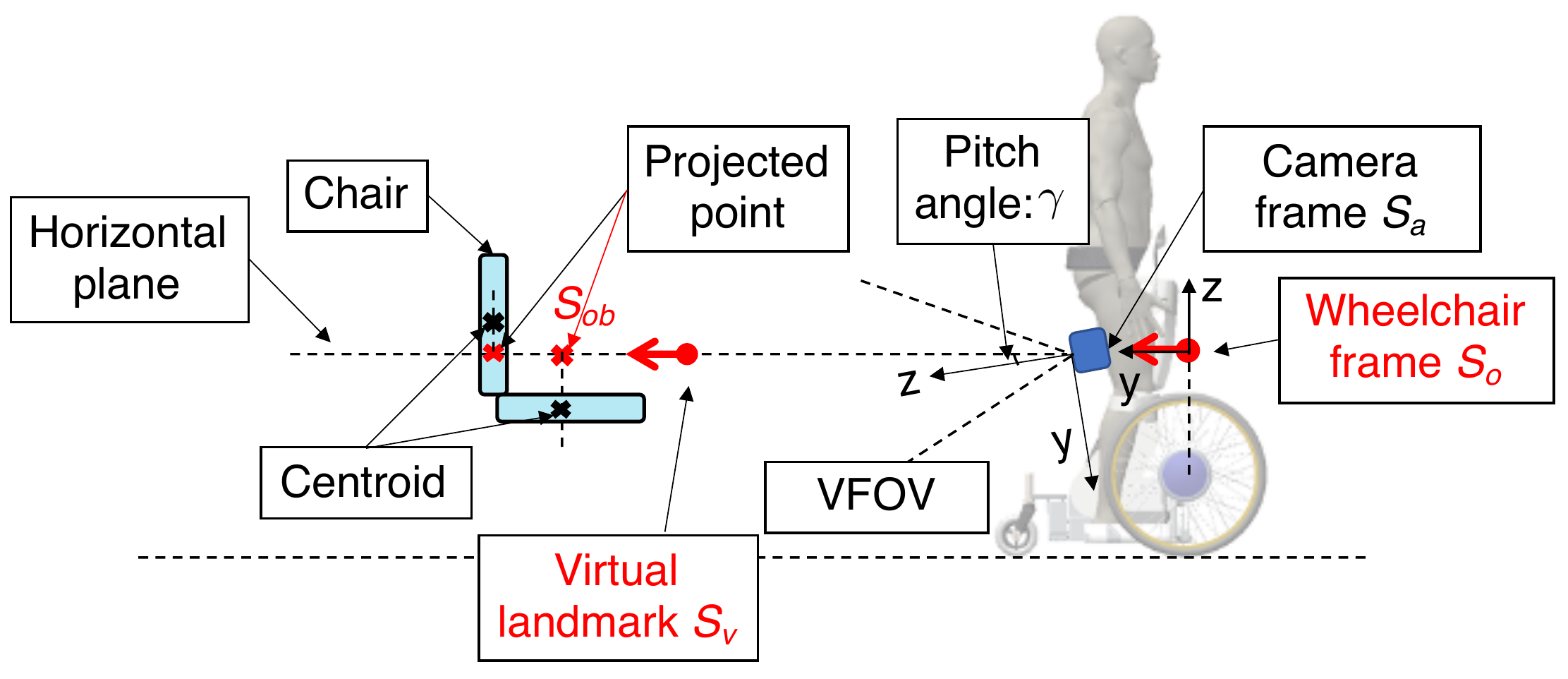}
\caption{Docking system example configuration: depicting a chair in blue with back and bottom centroids marked with black crosses, the virtual landmark and goal pose with red arrows, camera pitch angle as $\gamma$, one coordinate frame fixed at the sensing camera and another at the wheelchair.}
\label{co}
\end{center}
\end{figure}

For wheelchair docking control, a camera rigidly fixed on the wheelchair body is the most common and viable choice as source of visual feedback for generating the virtual landmark. Nonetheless, the main challenge comes from combining a standard powered wheelchair nonholonomic constraints and the FOV constraint of the camera into a single efficient solution for embedded implementation. 

Improving autonomous docking for wheelchair users have been tackled in some previous works, as the proposed Automated Transport and Retrieval System (ATRS) for a smart wheelchair docking to a lift platform of a car based on LiDAR information \cite{Gao2008}. \cite{Ren2012} proposed a bed docking control method for the wheelchairs consisting of path planning and line trajectory tracking. The work in \cite{Jain2014c} showed a novel method for autonomous detection of safe docking locations using 3D point cloud data without any visual marking or environment customization requirements. While in \cite{wei2012} the focus was on a 3D semantic map based-shared control for a smart wheelchair, achieving an object-related navigation task such as door passage or furniture docking.
Nonetheless, these approaches have not handled the challenge from the FOV constraint even they all use visual sensors.

\subsection{Problem Definition}\label{sec_pd}
\begin{figure}[t]
\begin{center}
\includegraphics[width=0.6\linewidth]{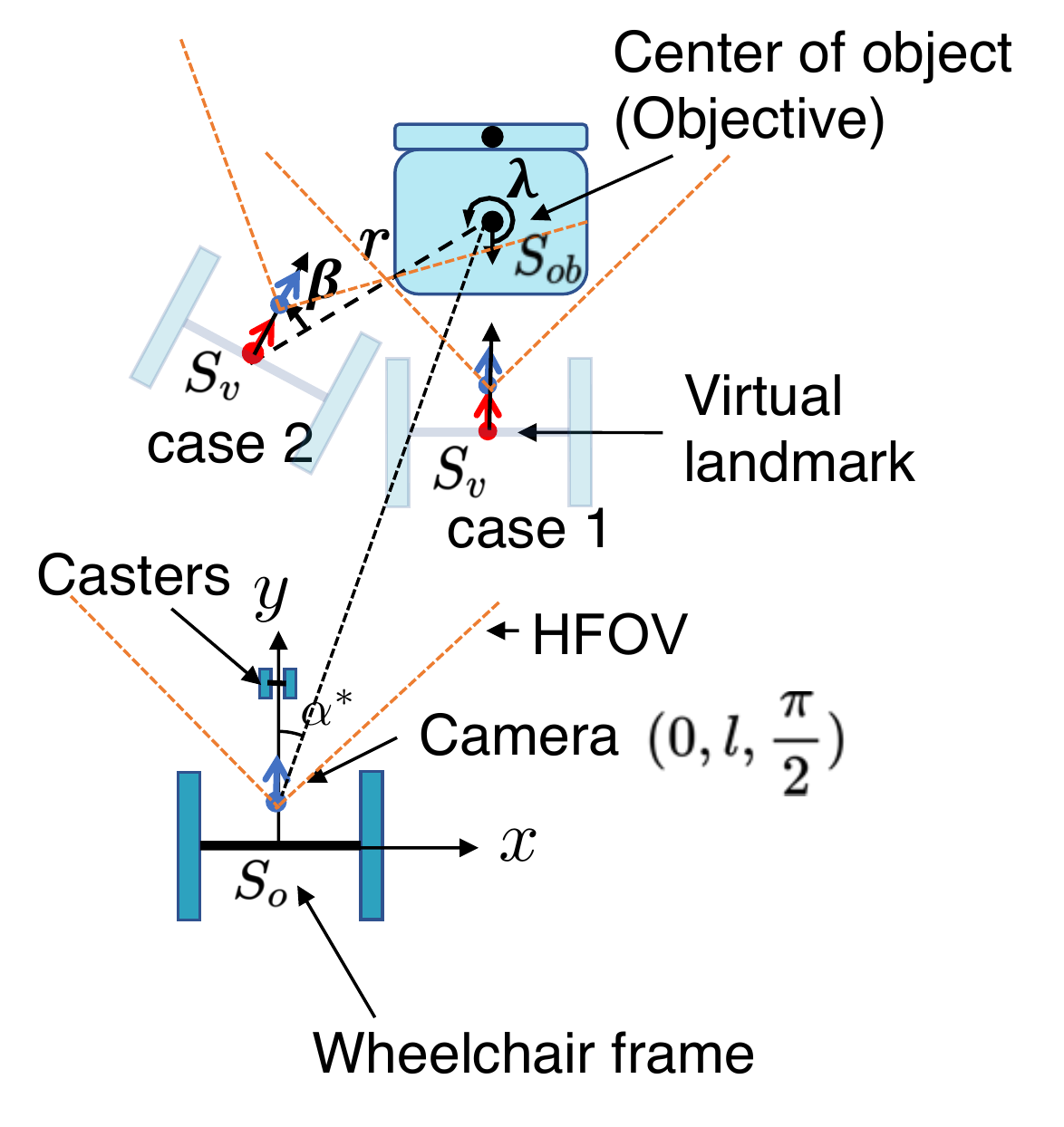}
\caption{Problem definition of wheelchair docking: case 1 is the case of docking to a pose which faces to the object, case 2 is the case of docking to the side of the object.}
\label{vl}
\end{center}
\end{figure}

The problem of keeping a real object inside camera's horizontal field of view (HFOV) and docking to a virtual landmark inferred from the same object is proposed here as described in Fig. \ref{vl}, as a novel description that tends to simplify the controller for embedded applications.

First, we set two polar coordinate systems on the object ($S_{ob})$ and the virtual landmark individually, where one pole lies on the center of the object, another pole lies on the virtual landmark, one polar axis coincidents with the centerline of the object, another polar axis coincidents with the posture of the virtual landmark. Then the relative pose between the virtual landmark and the object was expressed by $(r, \beta, \lambda)$, where $r$ indicates the distance between the objective and the virtual landmark, $\beta$ represents the angle between virtual landmark posture and the line connecting the Objective and the center of the robot, $\lambda$ indicates the angle between object centerline and the line connecting Objective and the center of the robot. 
On the other hand, the camera's location effect for constraining the control of the robots was defined by $(l, \bar{\alpha})$. For simplicity and in order to reduce the effect of the FOV constraint, we set the centerlines of the camera aligned with the robot, therefore, the pose is set to $(0, l, \frac{\pi}{2})$. Herewith, the angle $\bar{\alpha}$ should be a constant for the chosen camera and equals to half of the camera's FOV.

Our solution is to consider the object for creating a virtual target (landmark) as a reference frame, then the problem could be rephrased as a wheelchair with a camera fixed at $(0, l, \frac{\pi}{2})$ docks to pose $(r, \beta, \lambda)$ while keeping $\alpha^{*}$ within the FOV defined by $\bar{\alpha}$ along the trajectory, where $\alpha^{*}$ is the bearing angle between the centerline of the camera and the connecting line of camera and Objective. 
To the authors best knowledge there has not been such a complete framework that generalizes both the virtual landmark as $(r, \beta, \lambda)$ and camera parameter as $(l, \bar{\alpha})$, which is quite challenging for the controller design. Most of the literature consider either one or all of $(r, \beta, l)$ as zero which simplifies the problem, but in practices it is not always true, e.g., $l$ could be nonzero due to possible robot physical limitation, $r$ could be nonzero when the size of the wheelchair and object are relatively big, $\beta$ could be nonzero due to the user's requirement.

\subsection{Related Work}\label{sec_rw}
Considering similar kinematics of underactuated vehicles keeping an Objective inside an HFOV along the docking trajectory, several works have addressed parts of this problem. In  \cite{sans2016vision} the authors proposed to use a logarithmic spiral as a planned path which could keep the Objective with a constant bearing angle along the trajectory, however, they used an additional straight line as a second stage to enter the docking station, which resulted in the transition between logarithmic spiral and straight-line not smooth. In a subsequent work \cite{Sans-Muntadas2017c}, they extended the docking path planning by combining a Fermat and logarithmic spiral’s properties together keeping the Objective within the HFOV and ensuring a smooth transition until the goal. However, in this case, the distance to the goal $r$ is expected to be zero, thus the problem solution is not directly extendable to our described situation in Fig. \ref{vl}. 

The work in \cite{DilanAmarasinghe2005c, Murrieri2004} proposed hybrid (or switching) controllers, which provided an efficient method to overcome the nonholonomic constraints, however, they are designed for a mobile robot without a rider and the results presented do not meet velocity smoothness expected by an end-user. Other possible approach to solve the docking manoeuvre was developed in  \cite{Maniatopoulos2013d} through a dual-model predictive controller (MPC) scheme to maintain the visibility of the Objective and a dipolar vector field to guarantee the convergence. Similarly, \cite{ke2016visual} also proposed to use MPC to deal with the multi-physical constraints including FOV in a mobile robot visual servoing problem. Nonetheless, this becomes very computationally expensive for embedded implementation.
The study in \cite{Widyotriatmo2015b} used a barrier function as a Lyapunov function candidate to give the HFOV constraint, the designed control law showed asymptotic stability around the origin, however, they also considered the $r$ as zero.
A similar problem was targeted in \cite{bhattacharya2007optimal} and \cite{sans2019path} with proposed optimal paths to deal with the FOV constraint, however, the target task was limited to point navigation instead of pose to pose docking as we present in this work.

In contrast to previous works, we propose a novel autonomous docking support system consisting of a virtual landmark estimation algorithm and a nonlinear feedback controller. Although most of the literature above considered $l$ and $\beta$ as zero, we extend the problem with a direct solution. The virtual landmark estimation algorithm is based on point cloud information from a low-cost RGB-D camera and relies on the geometry feature of the object without specific environmental markers. The nonlinear feedback controller is designed under consideration of the camera's limited HFOV, thus, we find a feasible space where the stability of the controller is proved theoretically and validated with experiments of a robot successfully docking into a chair with the state variables converging and the chair kept inside HFOV at all times.

The paper continues as follows: Section \ref{sec_method} presents the virtual landmark estimation algorithm and the design of a nonlinear feedback controller. Section \ref{sec_sys} presents the system overview of the control architecture. Sections \ref{sec_eval} presents experimental setup and results in terms of convergency of the state variables and no violation of the HFOV constraint. Section \ref{sec_con} concludes this work and addresses future work.

\section{METHODOLOGY}\label{sec_method}
An autonomous docking support system is proposed to help the user approach a chair, bed, or other resting surfaces naturally and safely. 
The autonomous docking system consists of two main parts, the first part is the virtual landmark estimation, the second part is the nonlinear feedback controller. 
With a user riding on the device, we assume that at the moment of switching from user control to autonomous docking, the user aligned the camera HFOV, a detailed description of the feasible solution space is given in the next section. 

\subsection{Virtual Landmark Estimation}\label{pose_est}
We designed a virtual landmark in order to execute the docking without altering the environment which would limit the applicability of the system. The sensor should be able to recognize and estimate the pose of the docking object effectively. In a wheelchair, the required sensing distance should be from 4m to 0.2m, which caters to the user's requirement. We selected an RGB-D camera for three reasons: 
\begin{enumerate}
  \item The user should be able to freely choose the desired docking place in the environment (chairs or benches), thus it is impossible to install sensors at the docking place, e.g., IR used for vacuum robots, or QR code as a landmark.
  \item In order to recognize the pose of the chair precisely, 3D information is important, although LiDAR are very precise, but they are very expensive. 
  \item Different resting furniture (chairs, benches and beds) share similar geometry features, which could be described as one surface parallel to the floor and another surface vertical to the floor. In order to make a simple algorithm and reduce the computation, we only use the depth information from the camera (currently focused on chairs).
\end{enumerate}

With this in mind, we have chosen for our robot an Intel RealSense D435 camera (Intel, Santa Clara, CA, USA) as source, whose HFOV is 85 degrees.

For designing the camera pose w.r.t robot frame $S_o$ ($yz$ plane shown in Fig. \ref{co}, $xy$ plane shown in Fig. \ref{vl}), we consider two main factors: 1. Avoiding obstruction by the robot or the human body. 2. Keep the object inside the camera's FOV. 
The camera pose w.r.t $xy$ plane of frame $S_o$ was depicted in Fig. \ref{vl}, which sets $(0, l, \frac{\pi}{2})$, considering the limited HFOV a smaller $l$ is preferred. In the case of $zy$ plane, the pose of the camera is $(z_a, l, \frac{\pi}{2}+\gamma)$, $z_a$ is better to be set as a bigger value, the pitch angle $\gamma$ is selected as the camera should be able to keep the front edge of the object inside the vertical field of view (VFOV) when the wheelchair arrives at the virtual landmark. In this work, $l$, $z_a$ and $\gamma$ are tuned by experimentally.

For the point cloud processing, first, we cut the point cloud using a pass-through filter, resulting in a cube container with the most probable volume of the chair. Then, we use a VoxelGrid filter to downsample the point cloud, these two steps help reduce the computation. After that, we use the Euclidean Cluster Extraction method to get the clusters of the point cloud, with a selection of the minimum and maximum size, we further reduce the number of unneeded clusters. Finally, as assumed that the chair is in the FOV, we select the closest cluster as the chair.

The virtual landmark was then derived from the centroid of the chair bottom and back. To extract the bottom and back of the chair, we simply divide the chair with a height threshold, which is related to the height of the chair bottom, most chairs will fall within a similar threshold. For the implementation of the algorithm, we rely on the PCL open library \cite{Rusu_ICRA2011_PCL}. 
Because the docking control occurs on 2D, we project the centroid of the chair's bottom and back to the horizontal plane \cite{GeometryinCV}. With the coordinate frame $S_a$ fixed on the camera we derive the transformation to the horizontal plane as: 
\begin{equation}
y + z\tan\gamma = 0, 
\label{equ: hori_plane}
\end{equation}
then the projection point on the horizontal plane is defined as:
\begin{gather}
\left \{\begin{aligned}
&x_p = x_{or} \\
&y_p = y_{or} - \frac{y_{or} + z_{or}\tan\gamma }{1 + \tan^{2}\gamma} \\
&z_p = z_{or} - \tan\gamma \frac{y_{or} + z_{or}\tan\gamma }{1 + \tan^{2}\gamma}
\label{equ:projection}
\end{aligned}
\right.
\end{gather}
where \((x_{or}, y_{or}, z_{or})\) is the camera given original point. 
With (\ref{equ:projection}) we projected \((x_{pm}, y_{pm}, z_{pm})\) and \((x_{pk}, y_{pk}, z_{pk})\) for the chair bottom and  back centroids individually.

Subsequently, the fixed transformation between camera frame $S_a$ and wheelchair frame $S_o$ allows to easily obtain these two points w.r.t wheelchair frame $S_o$ as $(x_{pm}^O, y_{pm}^O)$ and $(x_{pk}^O, y_{pk}^O)$.

Following the diagram of the configuration in Fig. \ref{co},  the virtual landmark \((x_v^{'}, y_v^{'})\) was generated by: 
\begin{gather}
\left \{\begin{aligned}
&x_v = x_{pm}^O + r\frac{x_{pm}^O - x_{pk}^O}{d_{mk}}\\
&y_v = y_{pm}^O + r\frac{y_{pm}^O - y_{pk}^O}{d_{mk}}
\label{equ:dest}
\end{aligned}
\right.
\end{gather}

\begin{eqnarray}
\left[\begin{array}
  {c}{x_{v}^{'}} \\ 
  {y_{v}^{'}} \end
  {array}\right]
  =
  \left[\begin{array}{ccc}
{\cos{\lambda}} & {-\sin{\lambda}} \\ 
  {\sin{\lambda}} & {\cos{\lambda}} 
  \end{array}\right]
  \left[\begin{array}
  {l}{x_v - x_{pm}^O} \\ 
  {y_v - y_{pm}^O}
\end{array}\right]
+
\left[\begin{array}
  {l}{x_{pm}^O} \\ 
  {y_{pm}^O}
  \label{equ:dest_lambda}
\end{array}\right]
\end{eqnarray}

where $(x_{v}, y_{v})$ indicates the virtual landmark for $\lambda = 0$, and $(x_{v}^{'}, y_{v}^{'})$ represents the general case for any given $\lambda$.
$(x_{pm}^O, y_{pm}^O)$ and $(x_{v}^{'}, y_{v}^{'})$ are renamed as the Objective \(D(x_d,y_d)\) and virtual landmark \(C(x_c,y_c)\) in Section \ref{mc} for easier representation.

\subsection{Nonlinear Feedback Controller}\label{mc}
With a closed-loop on the virtual landmark from the on-board sensor available in real-time, we propose a feedback controller without a prior definition of the path, as done in \cite{bhattacharya2007optimal,sans2019path}.

\paragraph{Kinematic Formulation for Docking}
The motion of the wheelchair can be represented by the so-called unicycle model with \(v\) linear  and \(w\) angular velocities at the center of the powered wheels. \(x\), \(y\) and \(\theta\) denote wheelchair's position and orientation in the global reference system. The kinematics is defined by:
\begin{gather}
\left \{\begin{aligned}
&\dot{x}=v \cos \theta \\ 
&\dot{y}=v \sin \theta \\ 
&\dot{\theta}= w
\label{eq:unicycle}
\end{aligned}
\right.
\end{gather}

\begin{figure}[t]
\begin{center}
\includegraphics[width=0.8\linewidth]{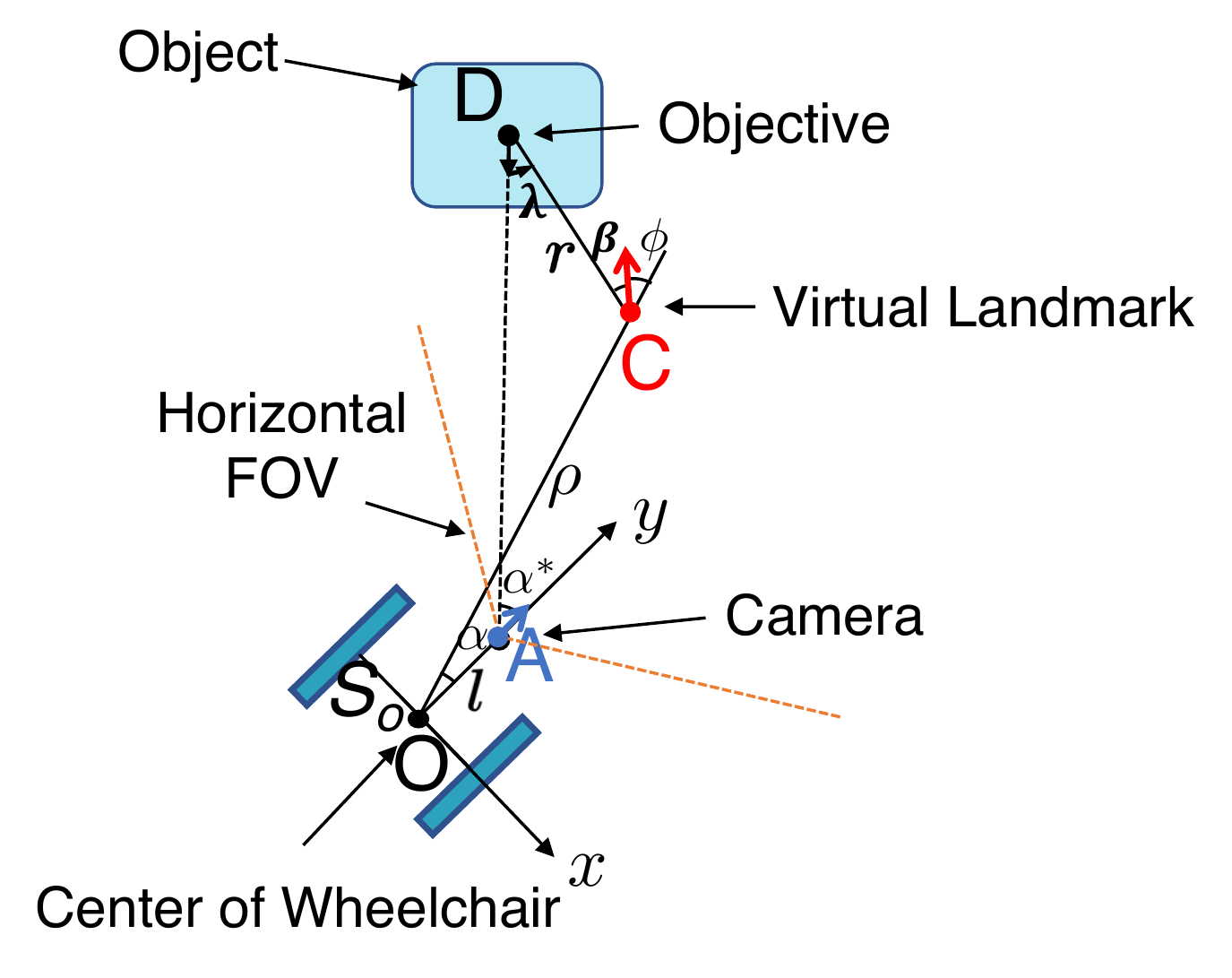}
\caption{Docking strategy, depicting the HFOV of the camera in yellow, the virtual landmark in red and the camera pose in blue, all points are w.r.t the wheelchair frame $S_o$.}
\label{ds}
\end{center}
\end{figure}

Fig. \ref{ds} depicts the wheelchair docking schematic detailing the coordinate system $S_o$ fixed at \(O(x_o,y_o)\) (the wheels center), \(A(x_a,y_a)\) indicates the location of the camera. \(r\) and \(l\) define the distance of CD and OA separately with ($r>l$). $\beta$ denotes the angle from CD to posture of virtual landmark, \(\rho\) denotes the length of OC, \(\alpha\) denotes the angle from OC to OA, \(\alpha^*\) describes the angle AD-OA, \(\phi\) denotes the angle from the goal pose of the virtual landmark to line OC defined as follows:
\begin{gather}
\left \{
\begin{aligned}
\rho = &\sqrt{(y_c - y_o)^2 + (x_c-x_o)^2} \\
\alpha = & \atantwo (y_c - y_o, x_c - x_o) - \\
& \atantwo(y_a - y_o, x_a - x_o) \\
\phi = &\atantwo(y_d - y_c, x_d - x_c) - \\
& \atantwo(y_c - y_o, x_c - x_o) - \beta
\label{rap}
\end{aligned}
\right.
\end{gather}

Inversely, point D could be expressed by \(\rho, \alpha, \phi\) as follows:

\begin{equation}
  \begin{cases}
    x_{d}= \rho\cos(\alpha + \frac{\pi}{2}) + r\cos(\alpha + \phi + \beta+ \frac{\pi}{2}) \\
    y_{d}= \rho\sin(\alpha + \frac{\pi}{2}) + r\sin(\alpha + \phi + \beta+ \frac{\pi}{2})
  \end{cases}
\end{equation}

The bearing angle \(\alpha^*\) could be then expressed as a function of $\rho, \alpha, \phi, l,r,\beta$, where $l,r,\beta$ are constants:
\begin{eqnarray}
\begin{aligned}
\alpha^* =& \atantwo(y_d - y_a , x_d - x_a) - \\
&  \atantwo(y_a - y_o, x_a - x_o) \\
 =& - \frac{\pi}{2} + \\
 & \atantwo(-l + \rho \cos\alpha + r \cos(\alpha + \phi + \beta), \\
 & -\rho \sin\alpha - r \sin(\alpha + \phi + \beta))
 \label{}
\end{aligned}
\end{eqnarray}

\paragraph{Control Law Design}
We propose a time-invariant control law that guarantees convergence and stability within the local conditions of the system. 
The kinematic equations of the nonholonomic unicycle robot from (\ref{eq:unicycle}) can be expressed using the state variables \(\rho, \alpha, \phi\) as follows:
\begin{gather}
\left \{
\begin{aligned}
&\dot{\rho} =-v \cos \alpha \\ 
&\dot{\alpha} =\frac{v}{\rho} \sin \alpha - w \\ 
& \dot{\phi} =-\frac{v}{\rho} \sin \alpha
\label{equ:kinematics}
\end{aligned}
\right.
\end{gather}
This formulation allows to control directly the relevant variables for constraining the FOV of the robot while docking.

Subsequently, we choose a Lyapunov function candidate $V = V_1 + V_2 +V_3 $ as :
\begin{gather}
\left\{\begin{array}{l}
V_{1}=\frac{\rho^{2}}{2} \\
V_{2}=\frac{\sin ^{2} \alpha}{2} \\
V_{3}=\frac{\phi^{2}}{2}
\end{array}\right.
\end{gather}
We design the controller by $v=k_{1} \rho \cos \alpha$, similar to the derivation in \cite{Widyotriatmo2015b}. Where $k_1$ is a first gain, and the linear velocity is proportional to the distance between the robot and virtual landmark, and inversely proportional to the angle $\alpha$, then the Lyapunov function becomes:
\begin{gather}
\left\{
\begin{aligned}
\dot{V}_{1}=&-k_{1} \rho^{2} \cos ^{2} \alpha \\
\dot{V}_{2}=&\sin \alpha \cos \alpha\left(k_{1} \sin \alpha \cos \alpha-\omega\right) \\
\dot{V}_{3}=&-k_{1} \sin \alpha \cos \alpha \phi \\
\dot{V}= &-k_{1} \rho^{2} \cos ^{2} \alpha+\sin \alpha \cos \alpha\left(k_{1} \sin \alpha \cos \alpha-\omega\right)\\
&-k_{1} \phi \sin \alpha \cos \alpha
\end {aligned}
\right.
\end{gather}

Finally, we designed $\omega$, first without the constraint of $\alpha^*$, we can have $\omega=k_{2} \sin \alpha \cos \alpha-k_{3} \phi$, in which case:
\begin{gather}
\begin{aligned}
\dot{V}= &-k_{1} \rho^{2} \cos ^{2} \alpha+\left(k_{1}-k_{2}\right) \sin ^{2} \alpha \cos ^{2} \alpha + \\
&(k_{3}-k_{1})\phi \sin\alpha \cos\alpha
\end{aligned}
\end{gather}

With $k_1 > 0$, $k_2 > k_1$ and $k_1 = k_3$,  we can guarantee $\dot{V}$ to be globally negative definite and global asymptotically stable. 
However, for the generalized case including $\alpha^*$, the solution is not straight forward. 
Using the formulation of $\alpha^*$ in (\ref{equ:alphaStar}), we adopted a constraint component $\sin ^{2} \bar{\alpha}-\sin ^{2}\alpha^{*}$ into (\ref{omega}) to regulate $\omega$, resulting in:
\begin{gather}
\omega=k_{2} \sin \alpha \cos \alpha-k_{3} \phi\left(\sin ^{2} \bar{\alpha}-\sin ^{2}\alpha^{*}\right)
\label{omega}
\end{gather}

Then,

\begin{gather}
\begin{aligned}
\dot{V}= & -k_{1} \rho^{2} \cos ^{2} \alpha+\left(k_{1}-k_{2}\right) \sin ^{2} \alpha \cos ^{2} \alpha \\
& -\left(k_{1}-k_{3}\left(\sin ^{2} \bar{\alpha}-\sin ^{2} \alpha^{*}\right)\right) \phi \sin \alpha \cos \alpha
\end{aligned}
\label{equ:v_dot}
\end{gather}

where we cannot guarantee that $\dot{V}$ is globally negative definite. However, we pursued a feasible space where $\dot{V}$ can be locally negative definite, thus making the controller stable within this region of initial states.
We achieved this solution by solving $(\rho, \alpha, \phi)$ through a numerical computation.

\paragraph{Designing the control gains by Lyapunov indirect method}\label{SV}
Before finding the feasible space to guarantee the stability, we can optimize the gains based on Lyapunov first method to guarantee the system asymptotic stability. With equations of $v$ and $\omega$ put into (\ref{equ:kinematics}) and the Jacobian linearization of $\dot{\rho}, \dot{\alpha}, \dot{\phi}$ with respect to $\rho, \alpha, \phi$ at the origin yields:

\resizebox{.47 \textwidth}{!}{
  \begin{minipage}{\linewidth}
\begin{equation}
\begin{aligned}
\left[\begin{array}{c}
\dot{\rho} \\
\dot{\alpha} \\
\dot{\phi}
\end{array}\right]=&\left[\begin{array}{ccc}
-k_{1} & 0 & 0 \\
0 & k_{1}-k_{2} & \frac{k_{3}\left(\sigma^{2} \sin ^{2} \bar{\alpha}-r^{2} \sin ^{2} \beta\right)}{\sigma^{2}} \\
0 & -k_{1} & 0
\end{array}\right] 
\left[\begin{array}{c}
\rho \\
\alpha \\
\phi
\end{array}\right]
\end{aligned}
\end{equation}
  \end{minipage}
}
  
Where $\sigma = \sqrt{r^2\sin^{2}{\beta} + (l - r\cos{\beta})^2}$, $\sin\alpha$ and $\sin\phi$ are assumed to be equivalent to $\alpha$ and $\phi$ individually when they are close to zero. 
The eigenvalues of the Jacobian matrix above are:
\begin{eqnarray}
\left \{\begin{aligned}
e_1 = &-k_1 \\ 
e_2 = &\frac{k_1-k_2+b}{2}\\ 
e_3 = &\frac{k_1-k_2-b}{2}
\label{equ:eigen}
\end{aligned}\right.
\end{eqnarray}

where $b = \sqrt{(k_1-k_2)^2\sigma^2 + 4k_1k_3(r^2\sin^2\beta - \sigma^2\sin^2\bar\alpha)}$. 
In order to keep all the eigenvalues of the Jacobian matrix negative, \(k_1\), \(k_2\), \(k_3\) are constrained as follows:
\begin{equation}
\left \{\begin{array}
{l}{k_{1}>0} \\ 
{k_{2}>k_{1}} \\ 
{k_{3}=\frac{(k_{1}-k_{2})^{2}\sigma^2}{4 k_{1} (\sigma^2\sin^2\bar\alpha - r^2\sin^2\beta)}}
\end{array}\right.
\end{equation}
With this, we can guarantee the system asymptotic stability. In order to keep the wheelchair under a low speed, here we choose \(k_1\), \(k_2\) as 0.15, 0.6 separately, \(k_3\) depends on the value of $l$, $r$ and $\beta$. 

\paragraph{Computation of feasible set}\label{FB}
With the designed control law and optimized $k_1, k_2, k_3$, we conduct a simulation in a given space $\Sigma \subset \mathbb{R}^{+} \times S^{2}$ of initial states: 
\begin{gather}
\left\{\begin{array}{l}
0 < \rho \leq 2.0 \\
|\alpha| \leq \frac{\pi}{3} \\
|\phi| \leq \frac{\pi}{3}
\end{array}\right.
\label{ise}
\end{gather}

\begin{figure}[t]
      \centering
  		\subfigure[]{\includegraphics[width=0.45\linewidth]{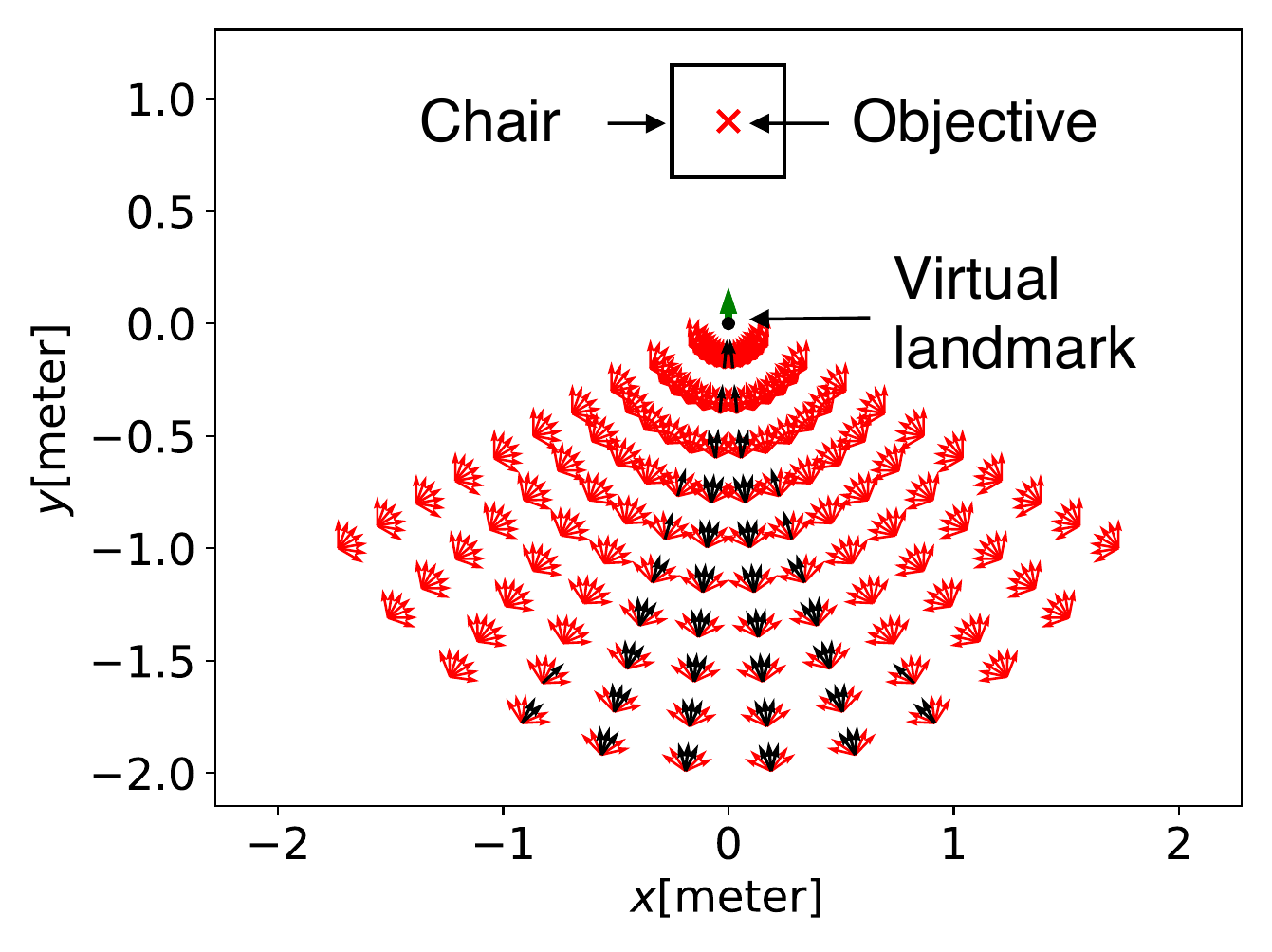}
		\label{}}
		\hfil
		\subfigure[]{\includegraphics[width=0.45\linewidth]{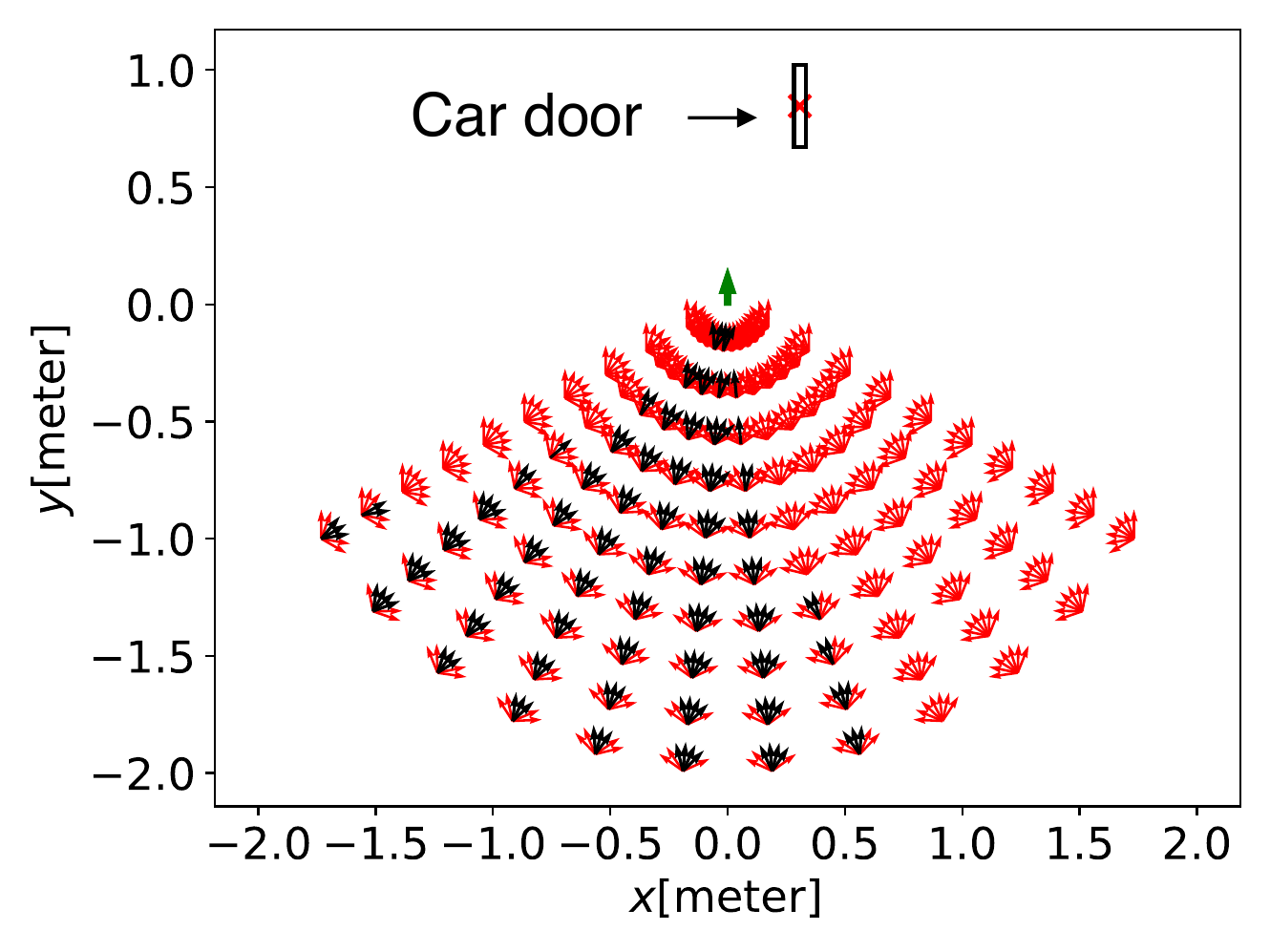}%
		\label{}}
      \caption{Example of the simulated initial states: with center of object (Objective) depicted with red cross, virtual landmark depicted with green arrow, the initial states which meet the requirement are depicted with black arrows, the initial states which do not meet the requirement are depicted with red arrows. (a) case 1: $(\lambda,r,\beta)$ = $(0^\circ,0.9m,0^\circ)$, $(l,\bar{\alpha})$ = $(0.26m,40^\circ)$, size of the object is $0.5m \times 0.5m$, (b) case 2: $(\lambda,r,\beta)$ = $(340^\circ,0.9m,-20^\circ)$, $(l,\bar{\alpha})$ = $(0.26m,40^\circ)$, size of the object is $0.05m \times 0.35m$.}
      \label{fs} 
  \end{figure}

\begin{figure}[t]
      \centering
  		\subfigure[]{\includegraphics[width=0.45\linewidth]{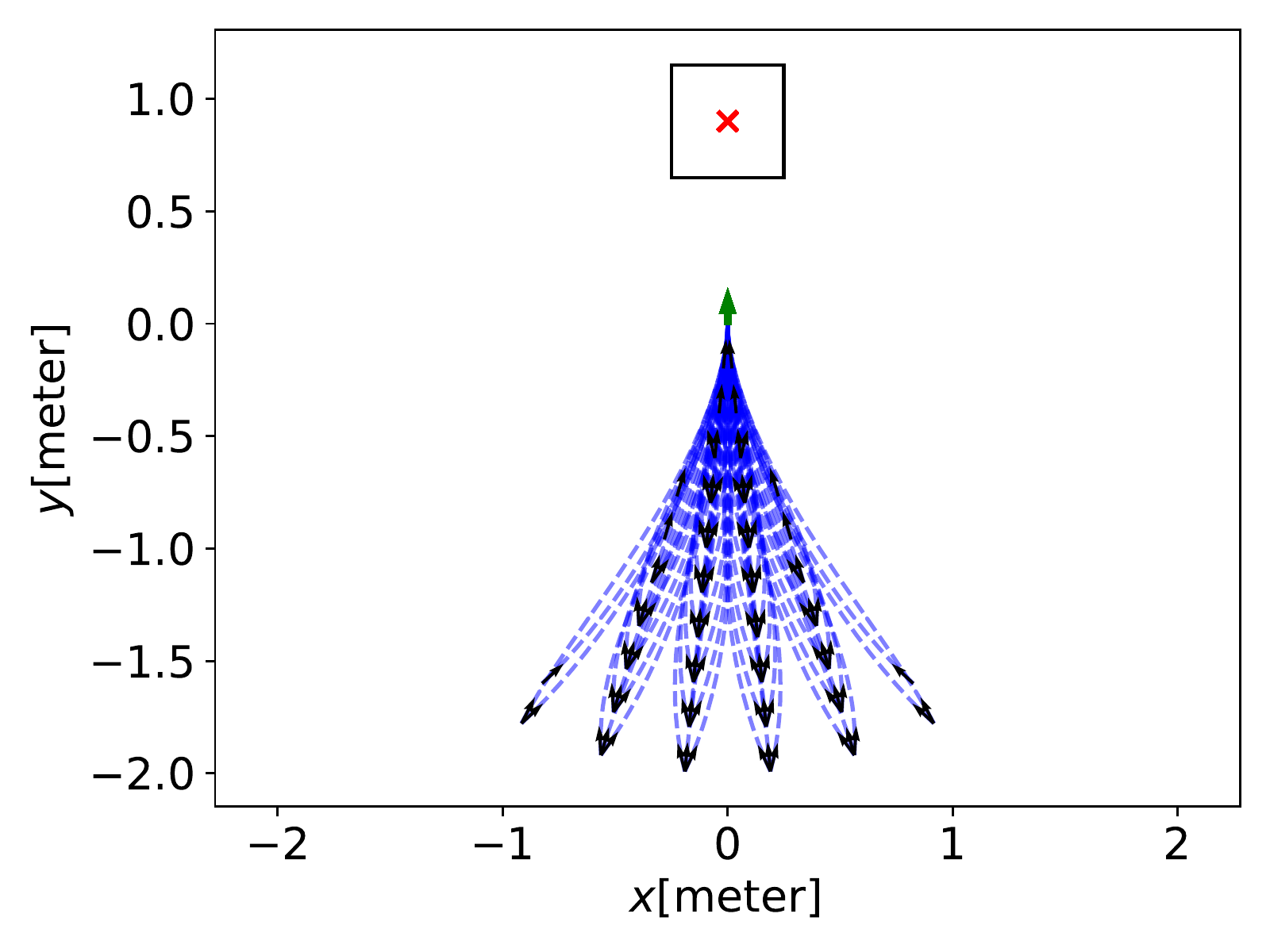}
		\label{}}
		\hfil
		\subfigure[]{\includegraphics[width=0.45\linewidth]{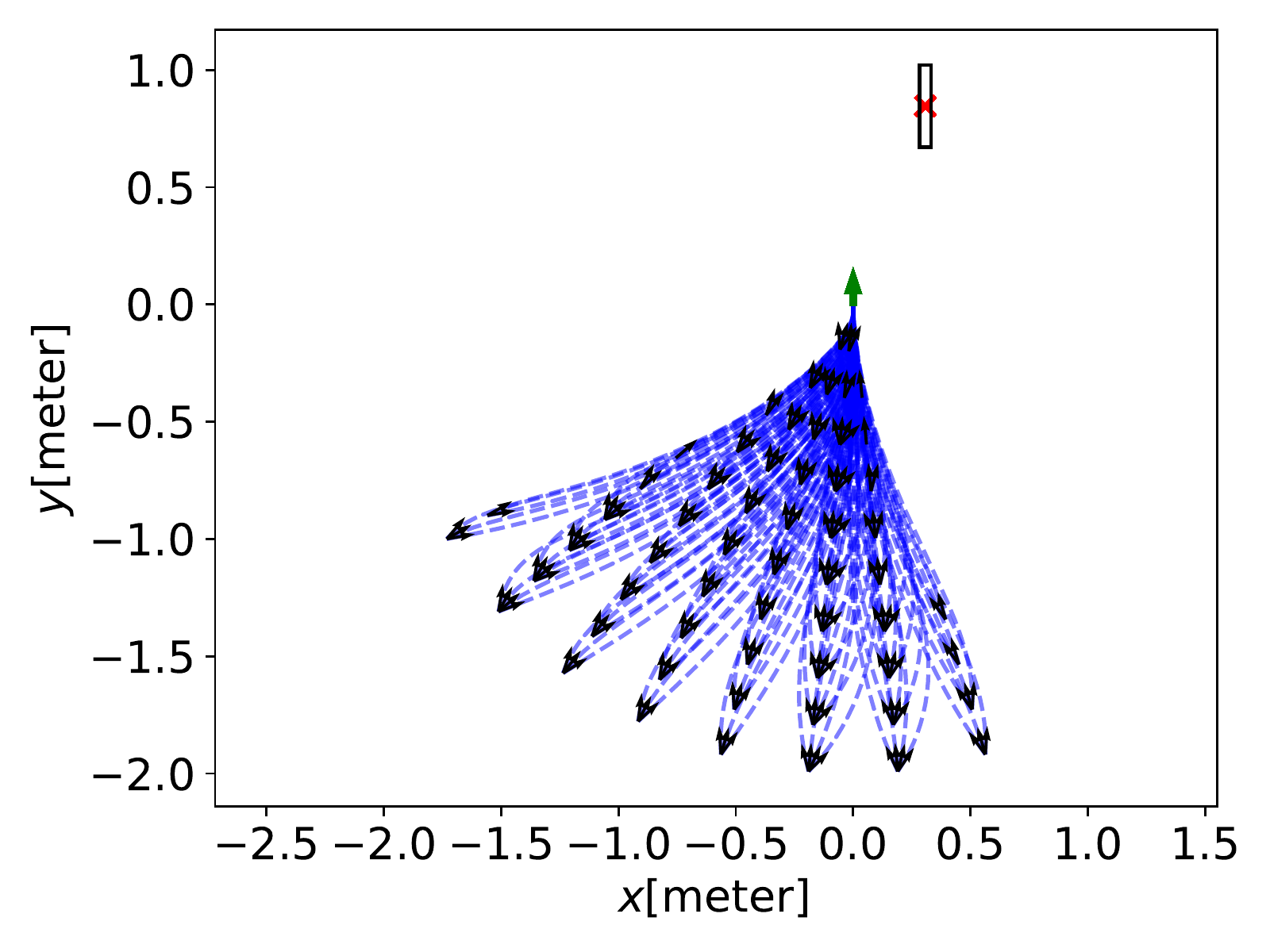}%
		\label{}}
      \caption{Trajectories of selected initial states: depicting with blue dotted lines the trajectory of wheelchair, depicting with black arrows the selected initial states. (a) case 1, (b) case 2.}
      \label{selected_trajectory} 
  \end{figure}

\begin{figure}[t]
      \centering
  		\subfigure[]{\includegraphics[width=0.45\linewidth]{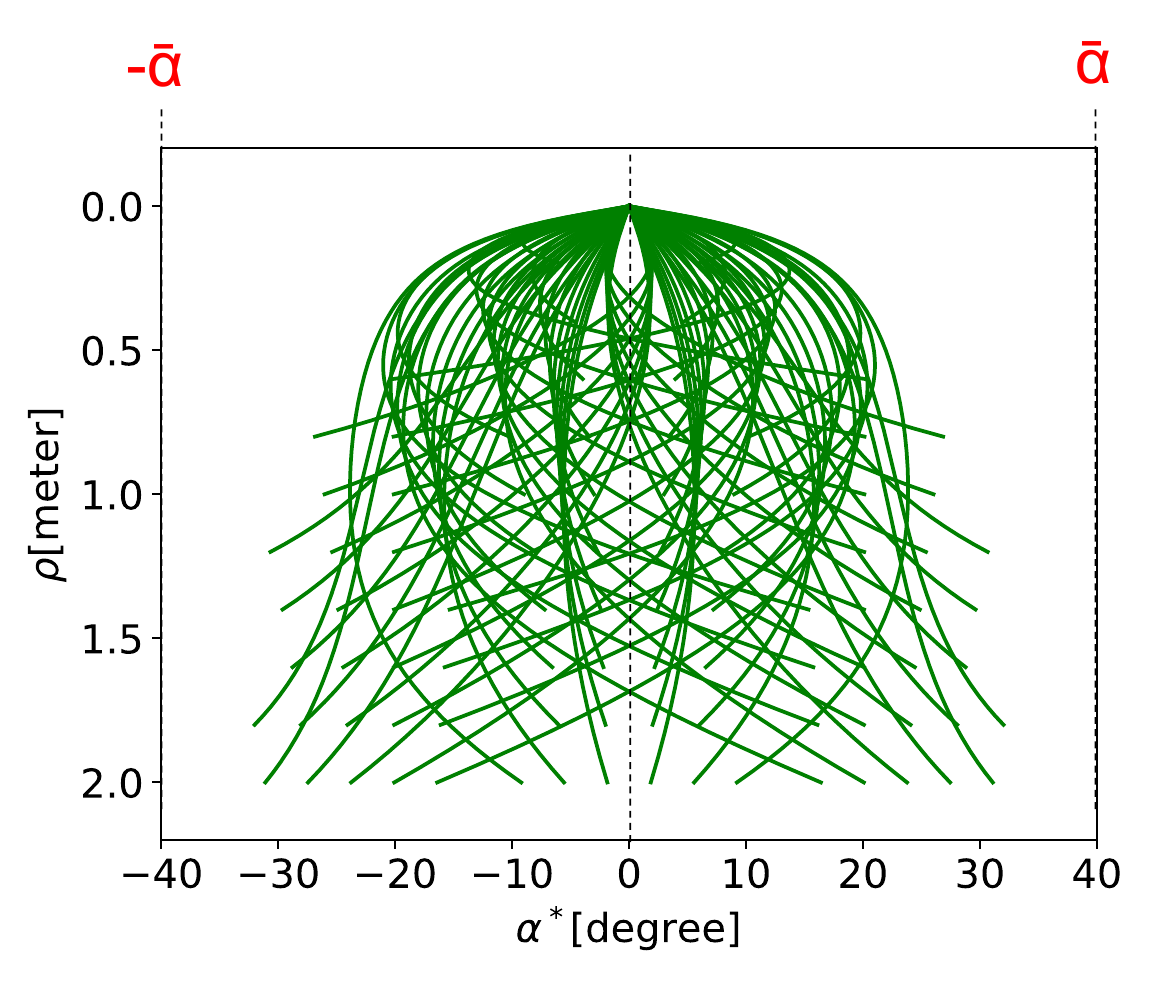}
		\label{}}
		\hfil
		\subfigure[]{\includegraphics[width=0.45\linewidth]{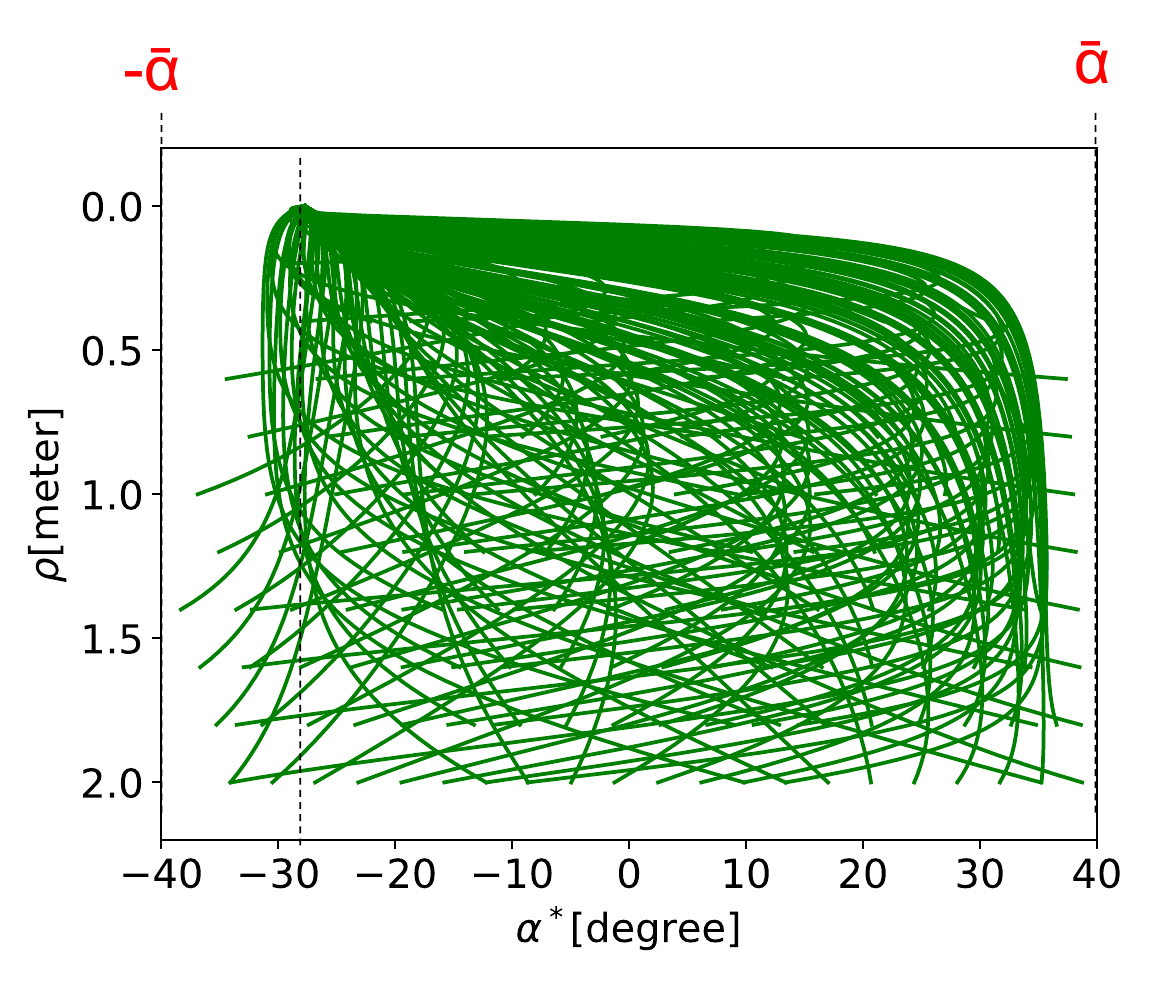}%
		\label{}}
      \caption{$\alpha^*-\rho$ of selected initial states. (a) case 1, (b) case 2.}
      \label{AlphaStar-Rho} 
  \end{figure}

The range of the space is based on the possible user requirement. We simulated for two examples: case 1: backward docking to a chair, case 2: wheelchair docking to a car laterally. The configuration of $(\lambda,r,\beta)$ and $(l, \bar{\alpha})$ differs. 
For readability of the figure, we only show part of the simulated results in Fig. \ref{fs}, we select the feasible set which meets requirements below: 1. The wheelchair converges to the virtual landmark. 2. Without violating HFOV constraint. 
For the second requirement, because the object is not a single point, in order to estimate the virtual landmark correctly, we need to keep the whole object including the edges inside HFOV. Those which meet the requirement are denoted by black arrows, those which do not meet the requirement are denoted by red arrows.
The simulation results in Fig. \ref{selected_trajectory} show the trajectories of the wheelchair, and Fig. \ref{AlphaStar-Rho} shows the relationship between \(\alpha^*\) and \(\rho\) highlighting the convergence to the desired pose from the found feasible subset of initial states.

Then we can fit the boundary of the feasible set with inequalities below:
\begin{gather}
\left\{\begin{array}{l}
\rho_{\min } \leq \rho \leq \rho_{\max } \\
|\phi + k_{4}\beta| \leq \frac{\rho}{k_{5}(r+l)} \\
|\alpha - k_{6}\phi| \leq  k_{7} \rho \left(1-\frac{|\phi+ k_{4}\beta|}{\left(\frac{\rho}{k_{5}(r+l)}\right)}\right) \bar{\alpha}
\end{array}\right.
\label{iequ}
\end{gather}

  \begin{figure}[t]
      \centering
		\subfigure[]{\includegraphics[width=0.45\linewidth]{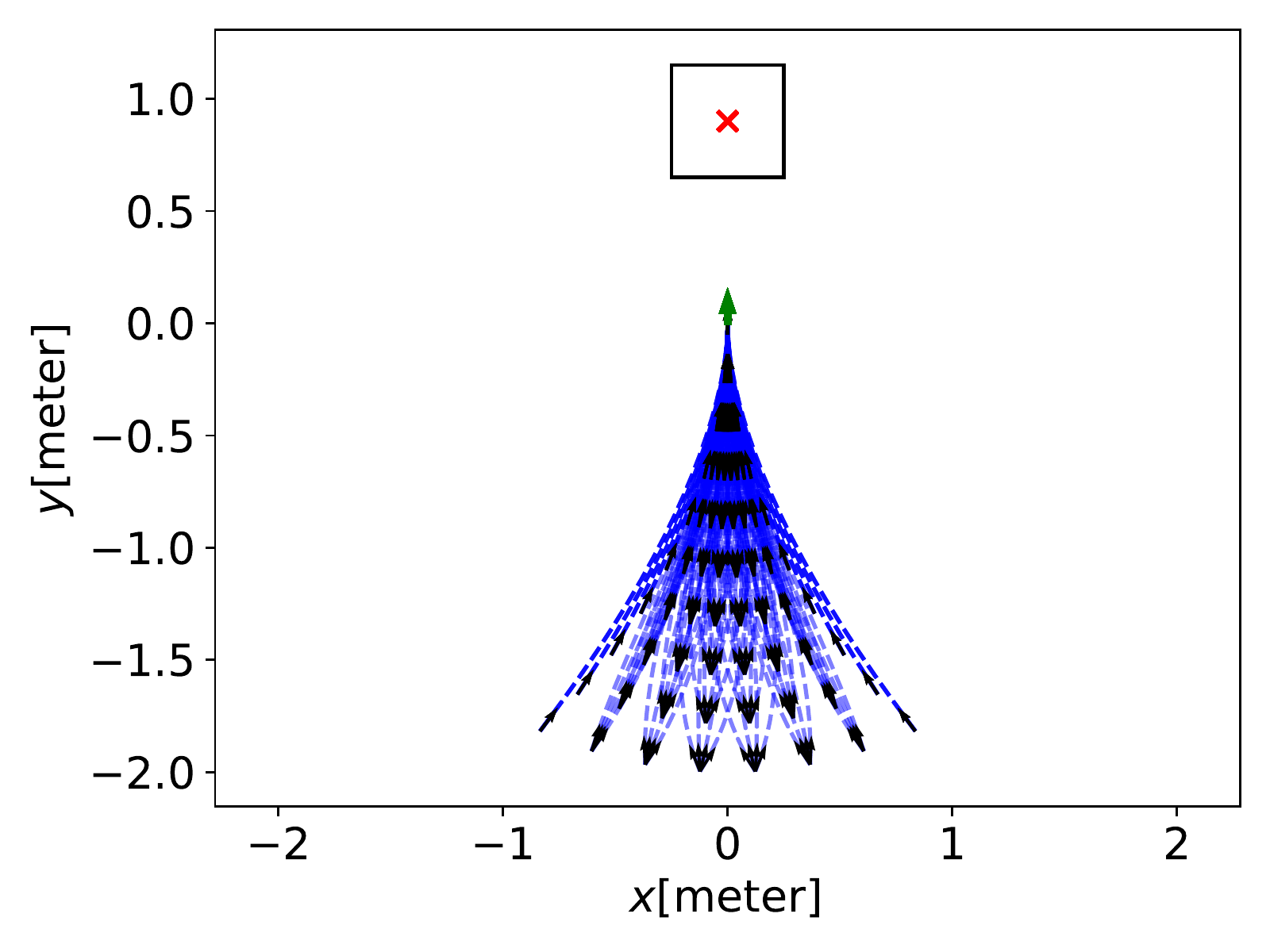}
		\label{Afitted_boundary_0_traj}}
		\hfil
		\subfigure[]{\includegraphics[width=0.45\linewidth]{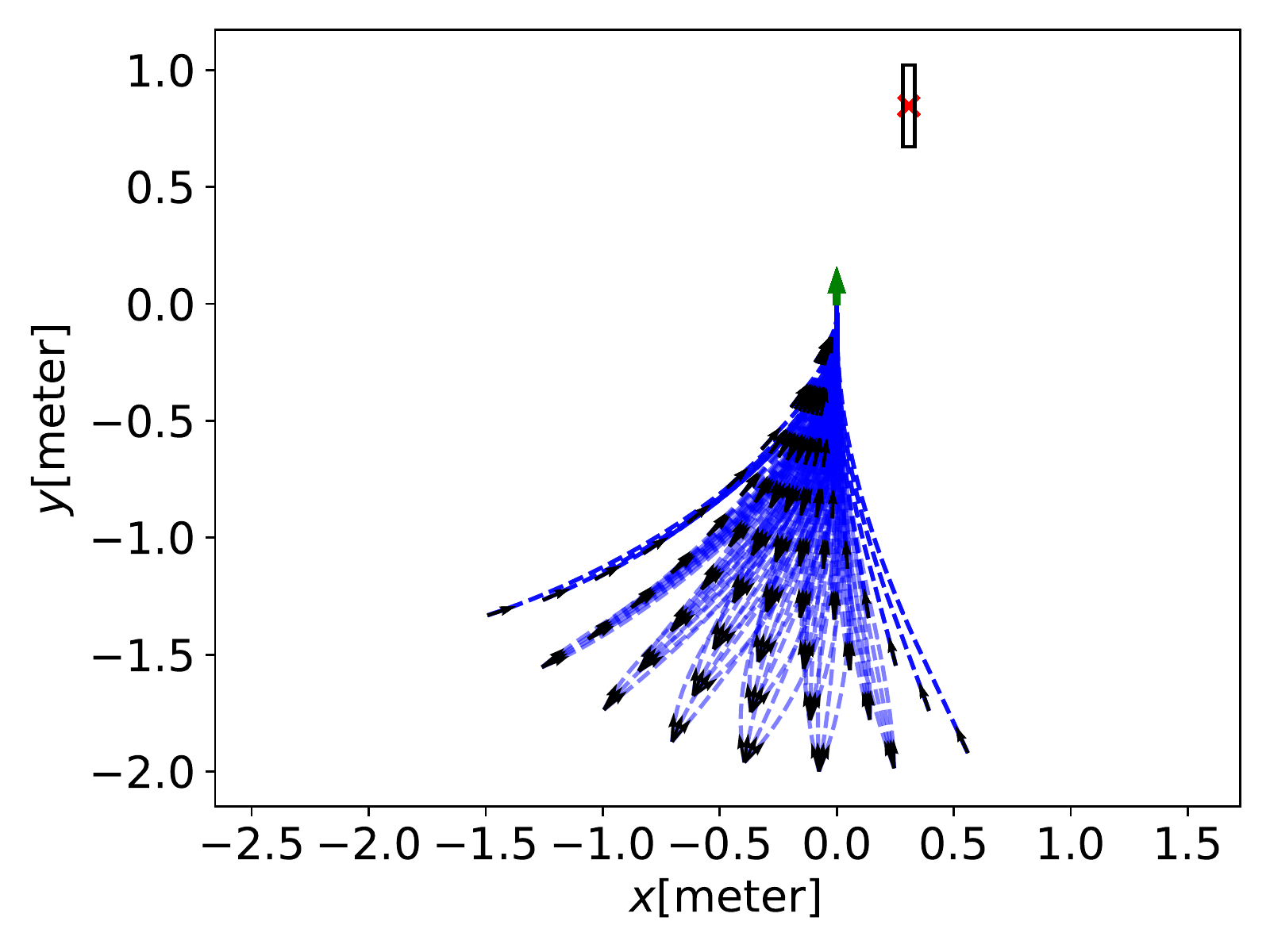}%
		\label{Afitted_boundary_beta_traj}}
      \caption{The example sets of fitted feasible space $F$ and trajectories. (a) case 1, (b) case 2.}
      \label{fitted_boundary} 
  \end{figure}
  
which is our ultimate feasible space $F \subset \mathbb{R}^{+} \times S^{2}$, where \(k_4\), $k_5$, $k_6$, $k_7$ are positive parameters that scale the shape of the space, which is also related to $(l,\bar{\alpha}, r,\beta)$, the example sets and trajectories of the ultimate feasible space are shown in Fig. \ref{fitted_boundary}.

We feed the fitted feasible space into $\dot{V}$ in (\ref{equ:v_dot}), validating that it is negatively definite in the feasible space. We also conduct the simulation with this fitted feasible space, all the trajectories could converge and $\alpha^*$ is also within the $\bar{\alpha}$. Thus we can say that the convergence and no violation of field of view are both guaranteed in this regulated control law.

\section{System Overview}\label{sec_sys}
The proposed system was implemented on the standing mobility vehicle Qolo \cite{Paez2018} through a Jetson Xavier (Nvidia Corporation) selected as a single-board computer. A custom-developed circuit was designed to communicate and drive the in-wheel wheelchair controller (YAMAHA Motor Corp., Iwata, Japan). Communicating with the mainboard through Ethernet protocol and to the wheels driver by CAN-bus communication, as built in a previous work in \cite{Leme2019}. 

\begin{figure}[t]
\begin{center}
\includegraphics[width=1.0\linewidth]{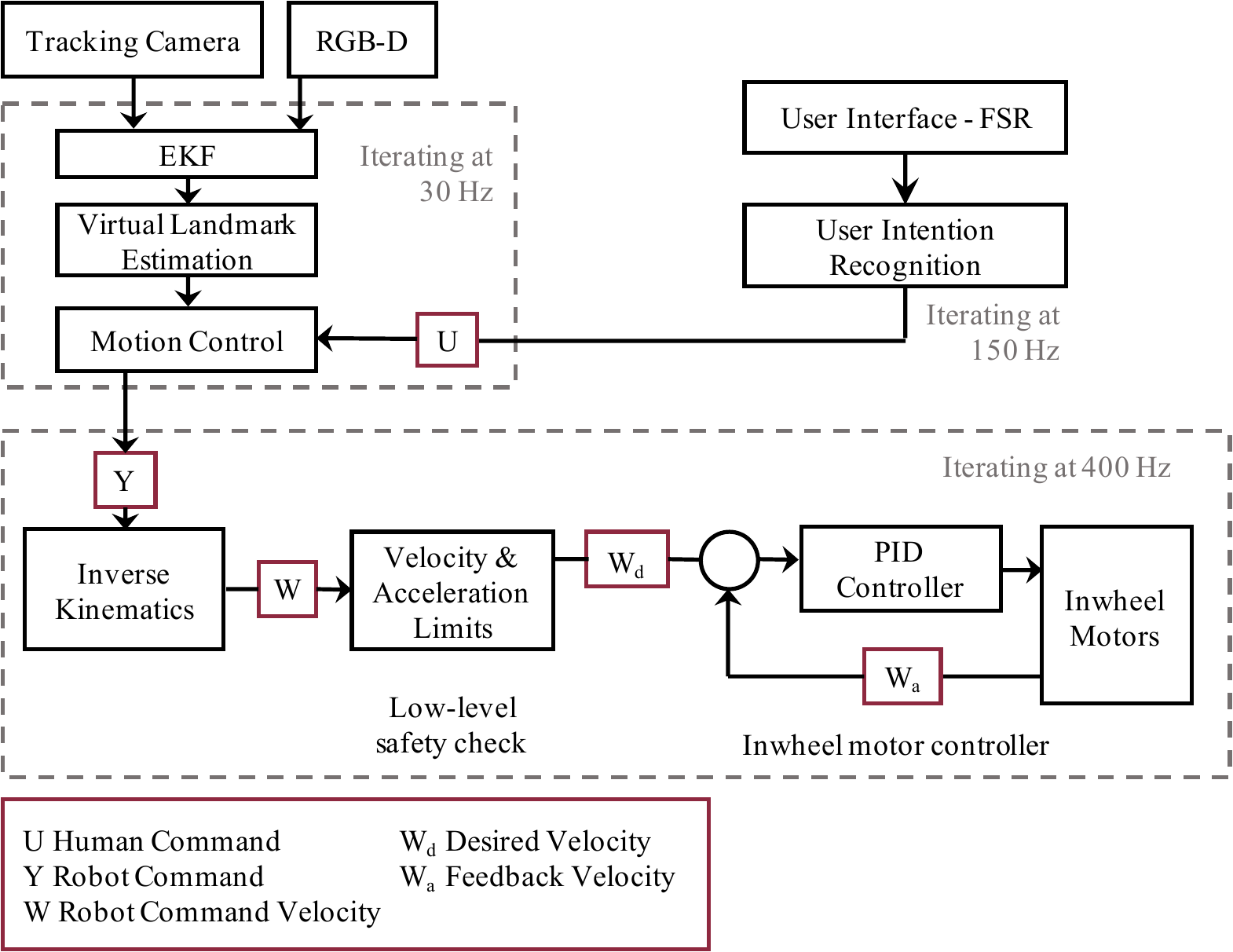}
\caption{Overall control system architecture for autonomous docking.}
\label{architecture}
\end{center}
\end{figure}

The overall control architecture is presented in Fig. \ref{architecture}. For safety, the autonomous docking control was designed to be supervised by the user. This initial interface was proposed in \cite{chen2020control} and the actuation method was explored in \cite{chen2020holistic}. Since our proposed docking procedure does not rely on manual control, we conducted a preliminary study without a user on-board. The overall system was implemented within the Robot Operation System (ROS).

\subsection{Extended Kalman Filter}\label{EKF}
The well-known Extended Kalman Filter (EKF) is used to attenuate the random noises in the pose estimation process. Here we fuse the centroid calculated by the camera D435 and the visual odometry information obtained from a camera T265 (Intel, Santa Clara, CA, USA).
The state variables used for filtering were chosen the Objective and virtual landmark w.r.t the wheelchair frame (Fig. \ref{ds}).

\section{Experiment Evaluation}\label{sec_eval}
\begin{figure}[t]
\begin{center}
\includegraphics[width=0.7\linewidth]{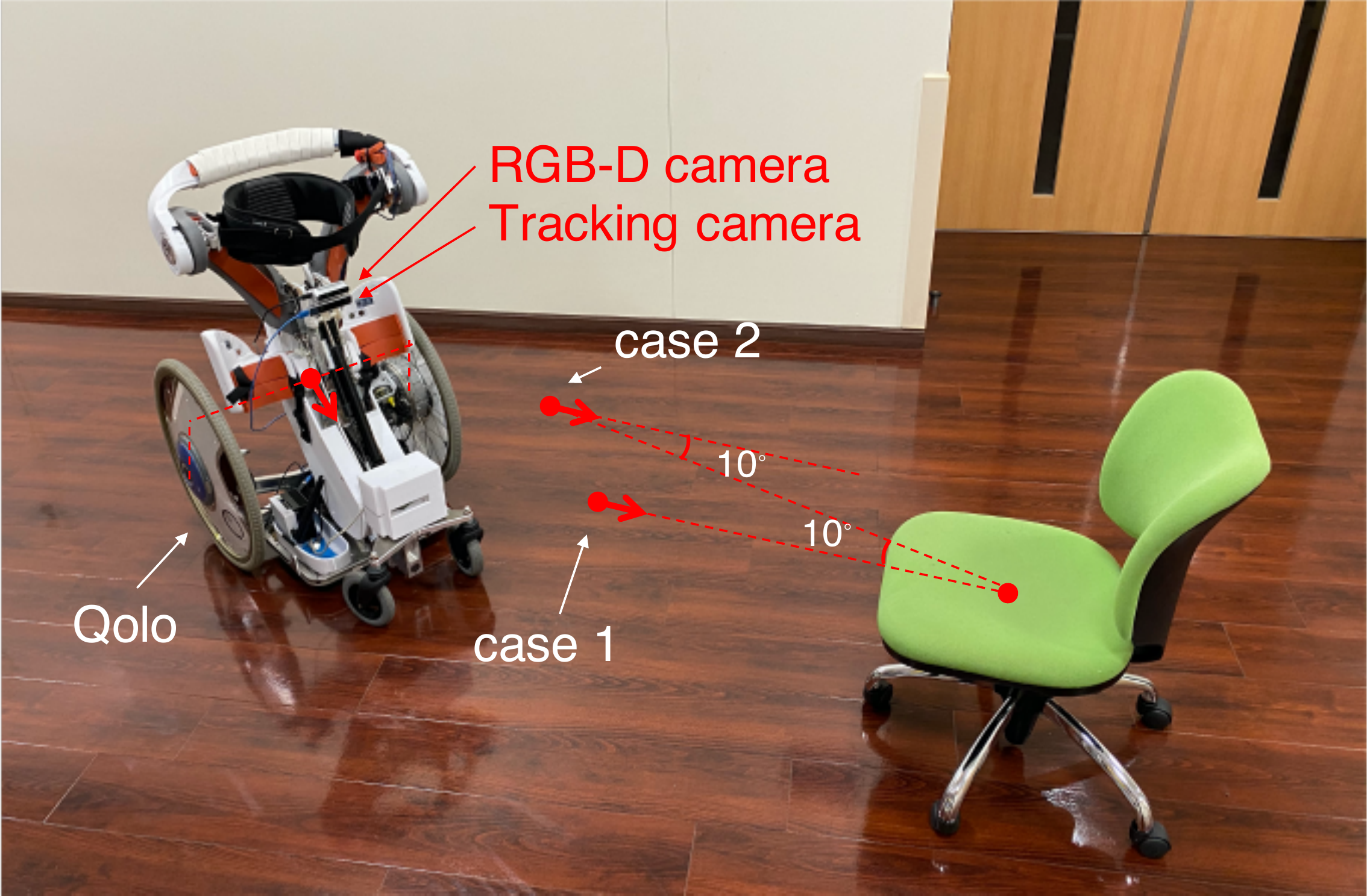}
\caption{Experimental evaluation snapshot of the robot Qolo and a desired docking object (a chair).}
\label{es}
\end{center}
\end{figure}

We conducted real-time experiments to verify the effectiveness of the proposed autonomous docking support system. The camera was fixed on the robot Qolo with $(l,\bar{\alpha})$ equal to $(0.26m,40^\circ)$. The goal object was a chair sized $0.5m \times 0.5m$. The initial conditions $(\lambda,r,\beta)$ were set to $(0^\circ,0.9m,0^\circ)$ and $(350^\circ, 1.1m, -10^\circ)$ for cases 1 and case 2 respectively. The experiment snapshot is shown in Fig. \ref{es}.

\subsection{Velocity Space Constraint}
In order to confirm the feasibility of the proposed control law, first, we investigated the feasible velocity space of our standing mobility device Qolo, in order to understand the dead-zone of the robot, i.e., the velocity space limited by the internal wheel velocity controller.

\begin{figure}[t]
\begin{center}
\includegraphics[width=0.6\linewidth]{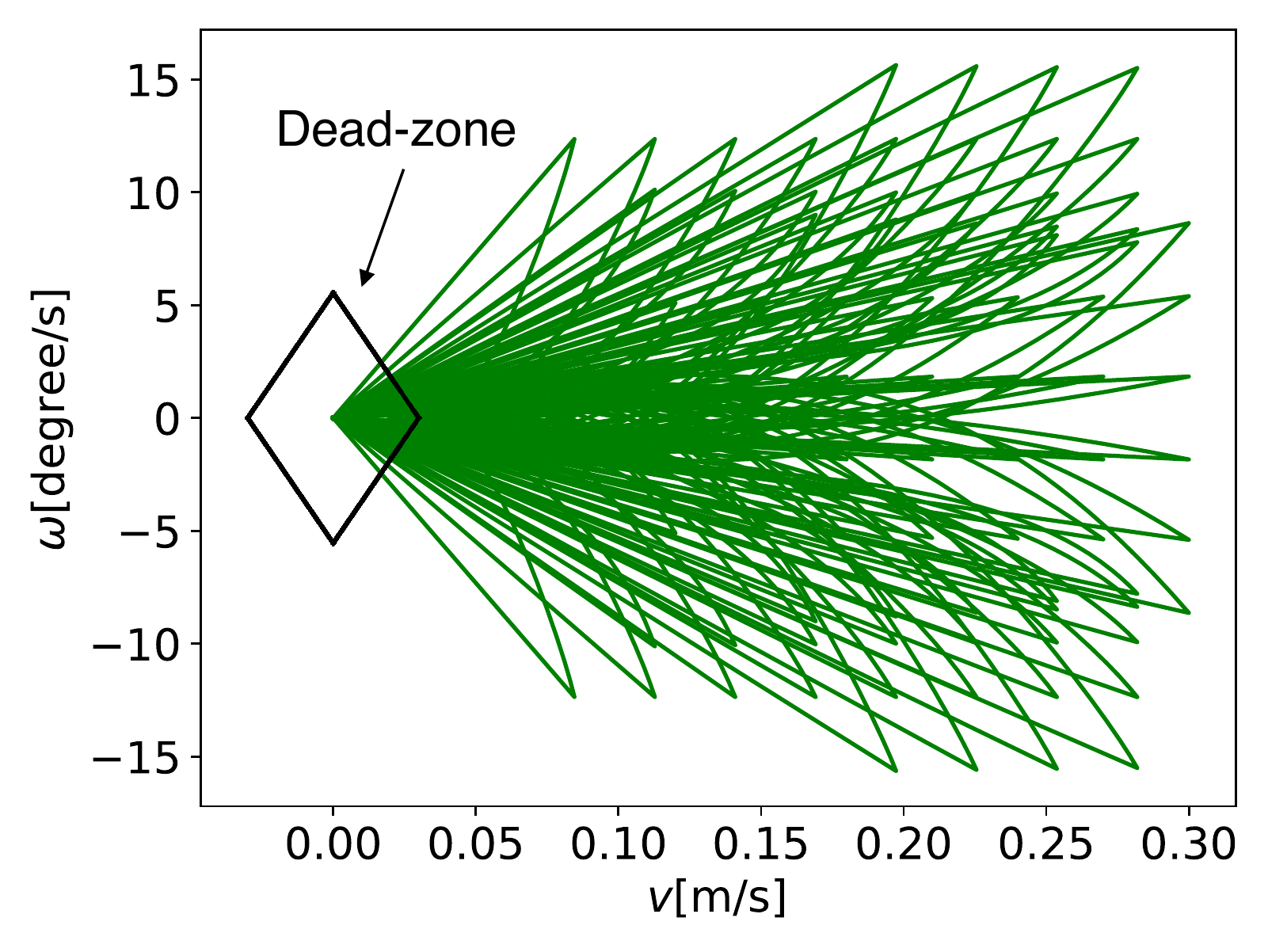}
\caption{Velocity Space analysis: dead-zone given by a test of the minimum velocity of the robot's wheels, in green the Simulated Velocity Space (SVS) of Qolo under the proposed control law (as simulated in Section \ref{mc}) .}
\label{FVS}
\end{center}
\vspace{-0.6cm}
\end{figure}

Here we assume that the left $v_l$ and right $v_r$ wheel velocities have the same independent limits: ${v_{l}, v_{r} \in \left(-v_{\max }, -v_{\min }\right) \cup \left(v_{\min }, v_{\max }\right)} $, where \(v_{min}\) and \(v_{max}\) denote the minimum and maximum velocity, then we can relate the robot's control through the kinematic model of a differential-drive robot \cite{siegwart2011}:
\begin{gather}
\left\{\begin{aligned} 
v &=\frac{v_{l}+v_{r}}{2}, 
\omega &=\frac{-v_{l}+v_{r}}{2 R} 
\end{aligned}\right.
\end{gather}
where $R$ denotes half of the distance between two wheels. 
As shown in Fig. \ref{FVS}, the Simulated Velocity Space (SVS) of Qolo under the proposed control law is denoted in green lines. In here, the dead-zone means the robot could not completely converge to the virtual landmark, therefore, for the assessment we evaluated the landmark goal within this area.

\subsection{Experimental Setup}\label{Experiment Setup}
An experiment was carried out in an indoor environment to verify the feasibility of the proposed autonomous docking system. Different initial states were tested to verify the real feasible space. As when Qolo gets closer to the chair, the oscillation of the virtual landmark will cause a bigger effect on the angular velocity, thus, we used a two-phase estimation strategy here, when Qolo was far away from the virtual landmark, we use the Extended Kalman Filter proposed in section \ref{EKF}, when Qolo gets over a threshold to the virtual landmark, we deactivated the virtual landmark estimation from depth camera side, and only use the tracking camera for odometry. 

As there exists a dead-zone of the wheelchair velocity, so the robot could not converge perfectly, what's more, we have not considered the dynamics of the robot, the effect from the casters, etc, so here we define a safety region $\Omega \subset \mathbb{R}^{+} \times S^{2}$ of \((\rho < 0.15m, |\alpha| < 10^\circ, |\phi| < 10^\circ)\), only when the robot enters this region, we consider it as a successful docking. Practicality speaking, this small deviation should be acceptable for the user, who could adjust any final alignment.

With this in mind, we tested the system with different initial states, as shown in Fig. \ref{fig:traj}. 

\subsection{Virtual Landmark Estimation}
\begin{figure}[t]
\begin{center}
        \subfigure[case 1]{\includegraphics[width=0.7\linewidth]{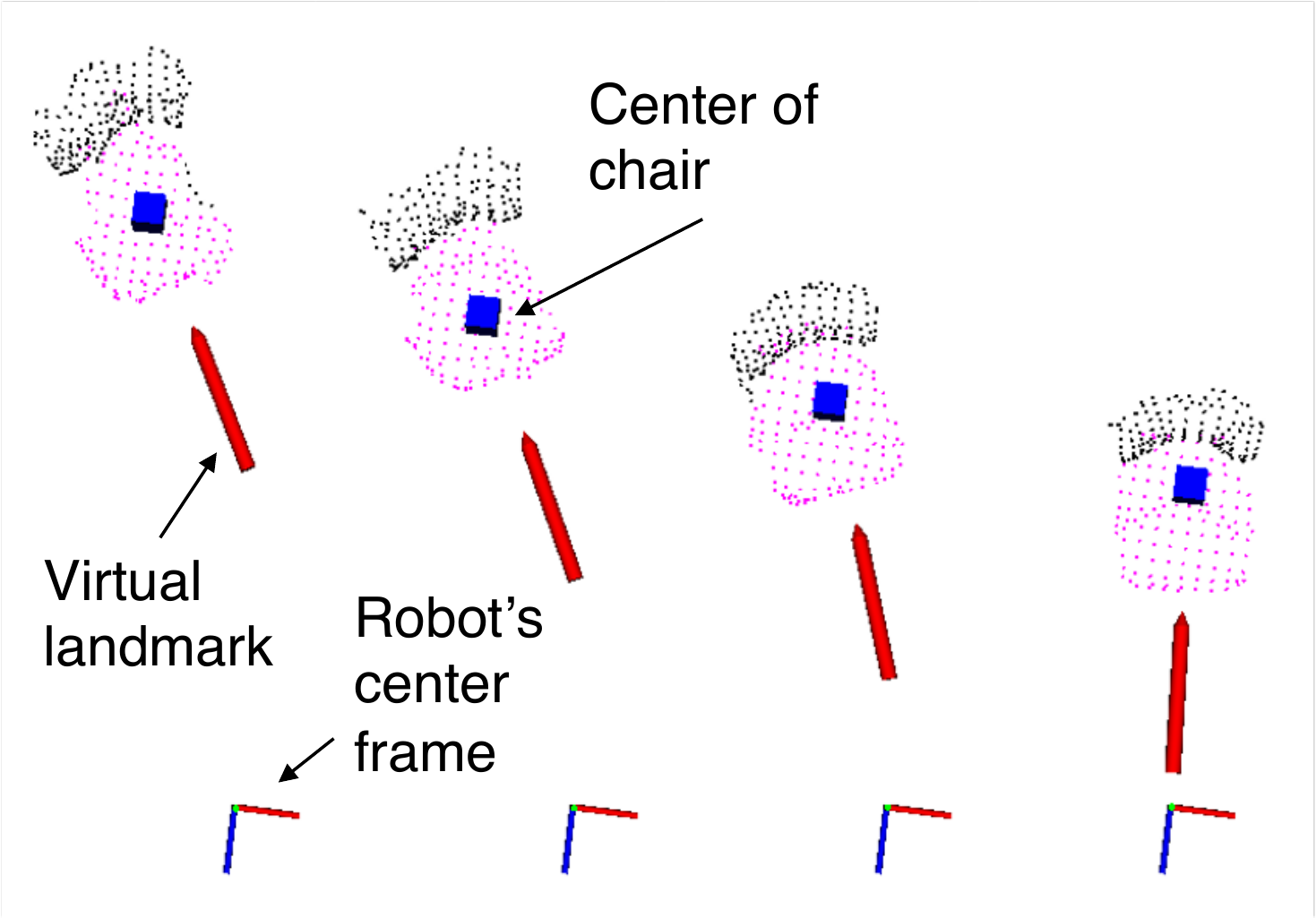}
		\label{fig:ae_case1}}
		\hfil
		\subfigure[case 2]{\includegraphics[width=0.7\linewidth]{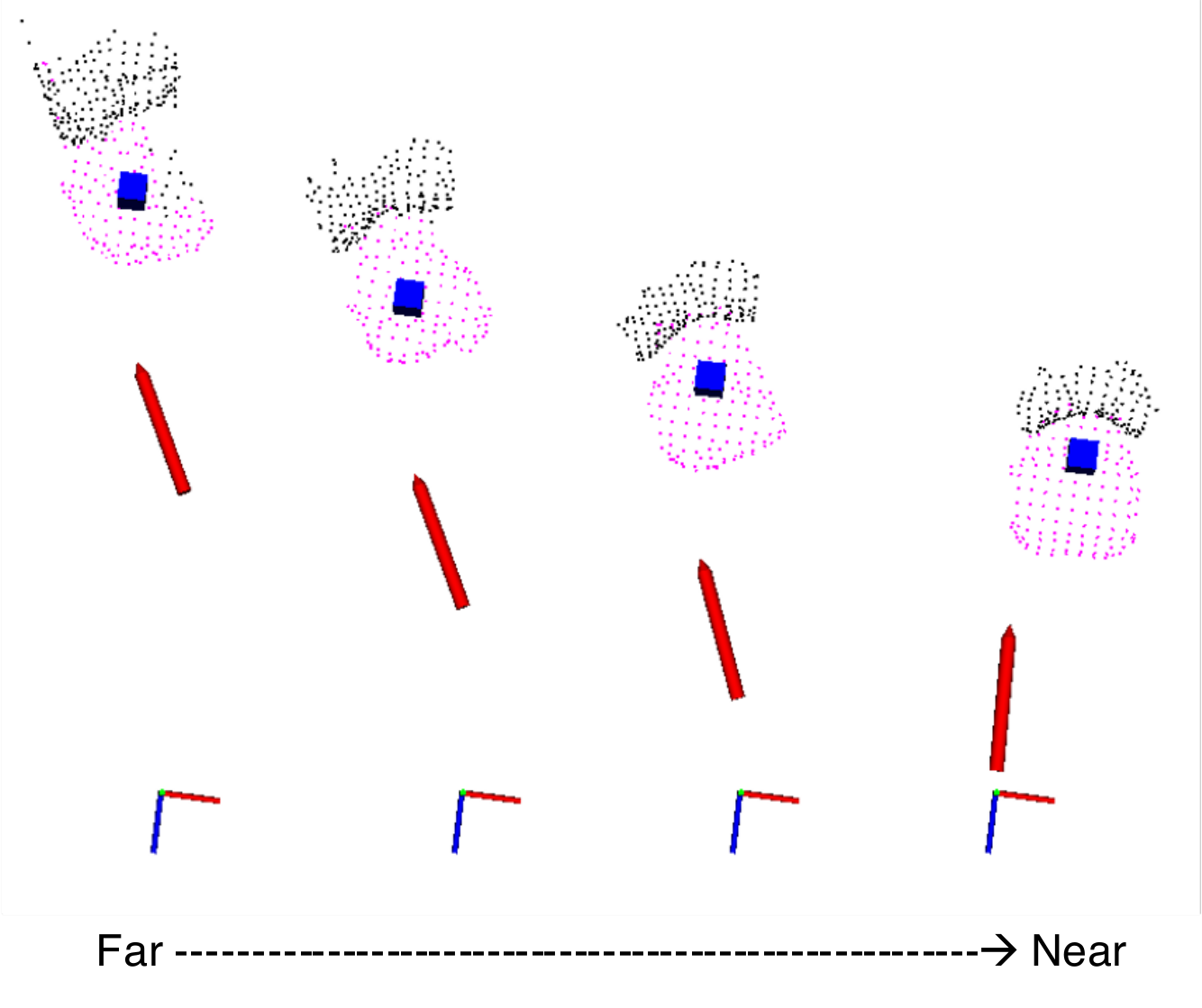}%
		\label{fig:ae_case2}}
\caption{Examples of virtual landmark estimation from different poses: black cluster denotes the chair back, the red cluster the chair bottom, the blue cube represents the center of chair, and the red arrow the virtual landmark.}
\label{ae}
\end{center}
\end{figure}

The virtual landmark estimation algorithm was validated by locating the robot at different relative poses to the chair (snapshots are shown in Fig. \ref{ae}). Qualitatively, the chair is clearly recognized from the environment, bottom and back are separated well. From left to right, the robot gets closer to the chair, as we can see, the accuracy of the estimated virtual landmark decreases with the relative distance.

\subsection{Convergence Confirmation}
\begin{figure}[t]
      \centering
  		\subfigure[Distance to the landmark $\rho$]{\includegraphics[width=0.45\linewidth]{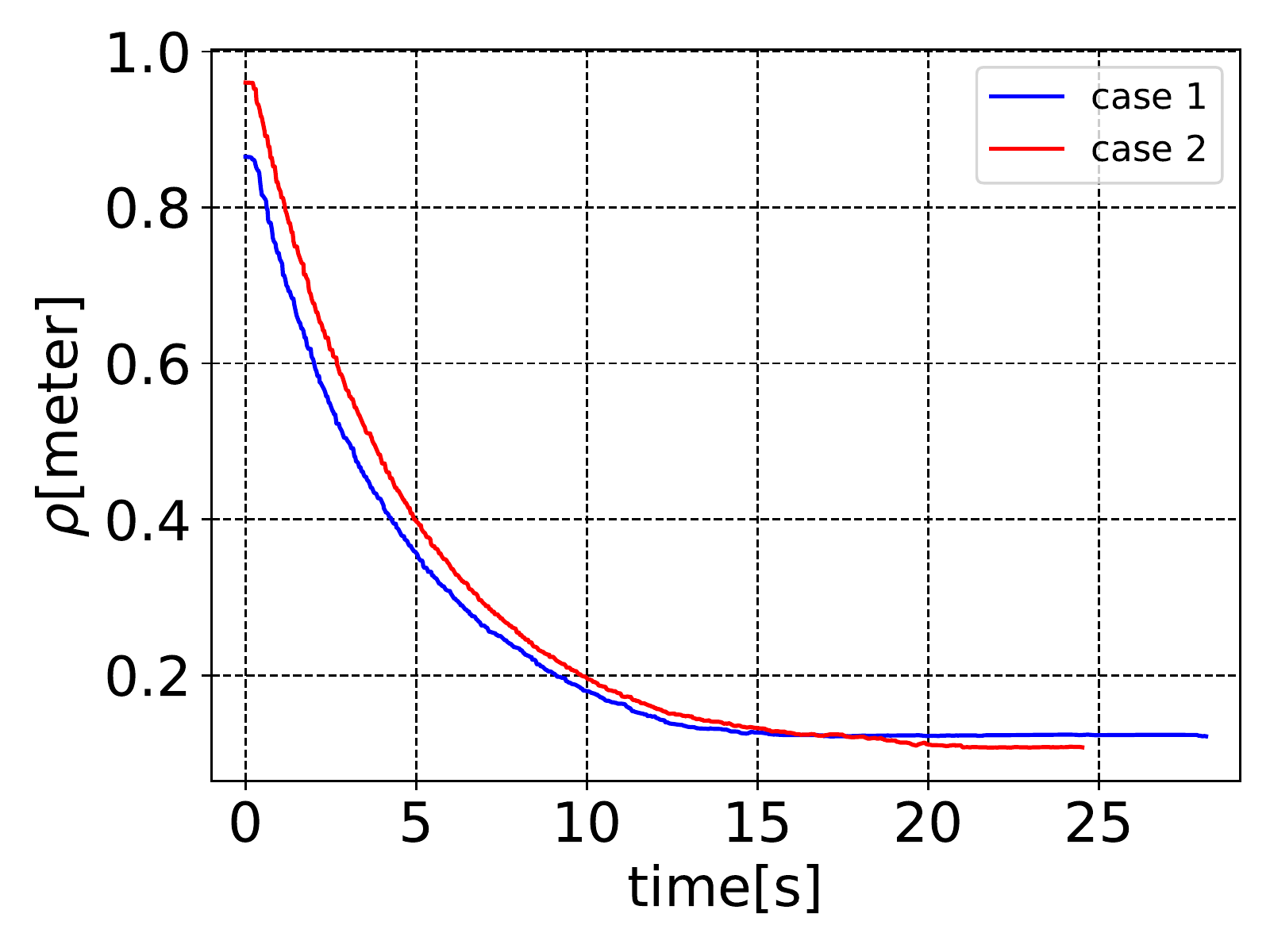}
		\label{fig:Rho}}
		\hfil
		\subfigure[Angle to the landmark  $\alpha$]{\includegraphics[width=0.45\linewidth]{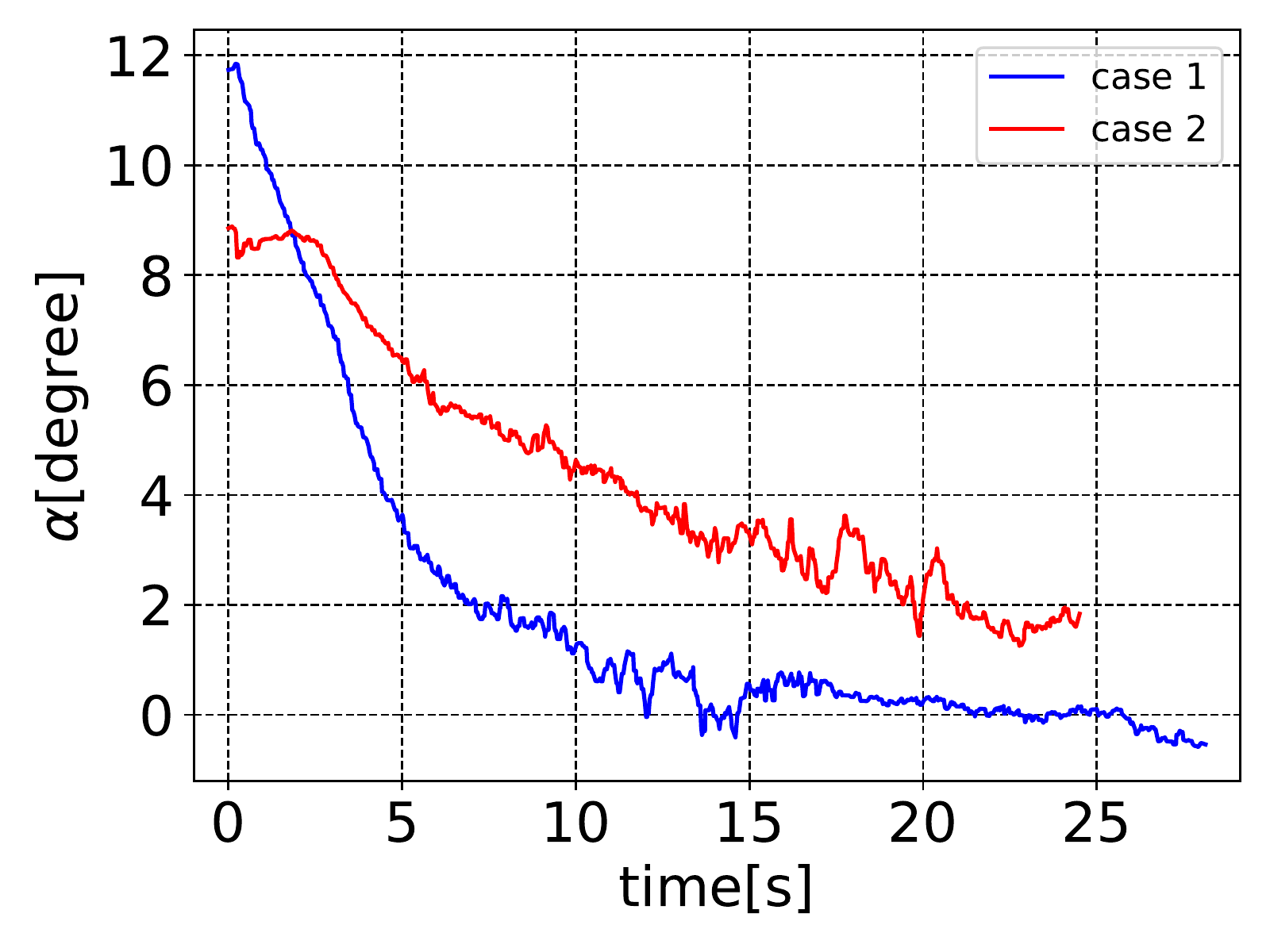}%
		\label{fig:Alpha}}
        \hfil
        \subfigure[Approaching angle to final pose $\phi$]{\includegraphics[width=0.45\linewidth]{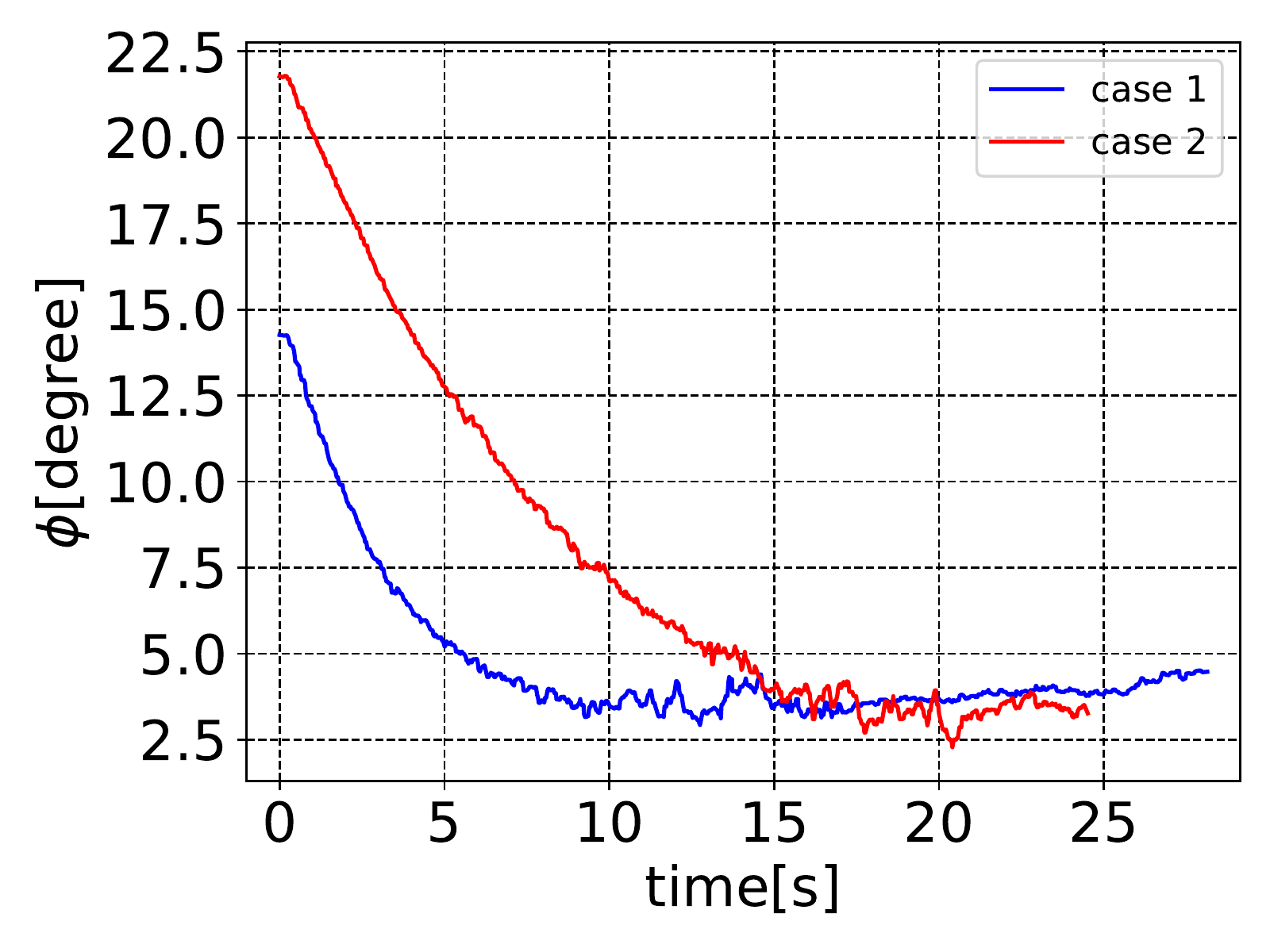}%
		\label{fig:Beta}}
        \hfil
		\subfigure[Sensor angle to the objective $\alpha^*$ ]{\includegraphics[width=0.45\linewidth]{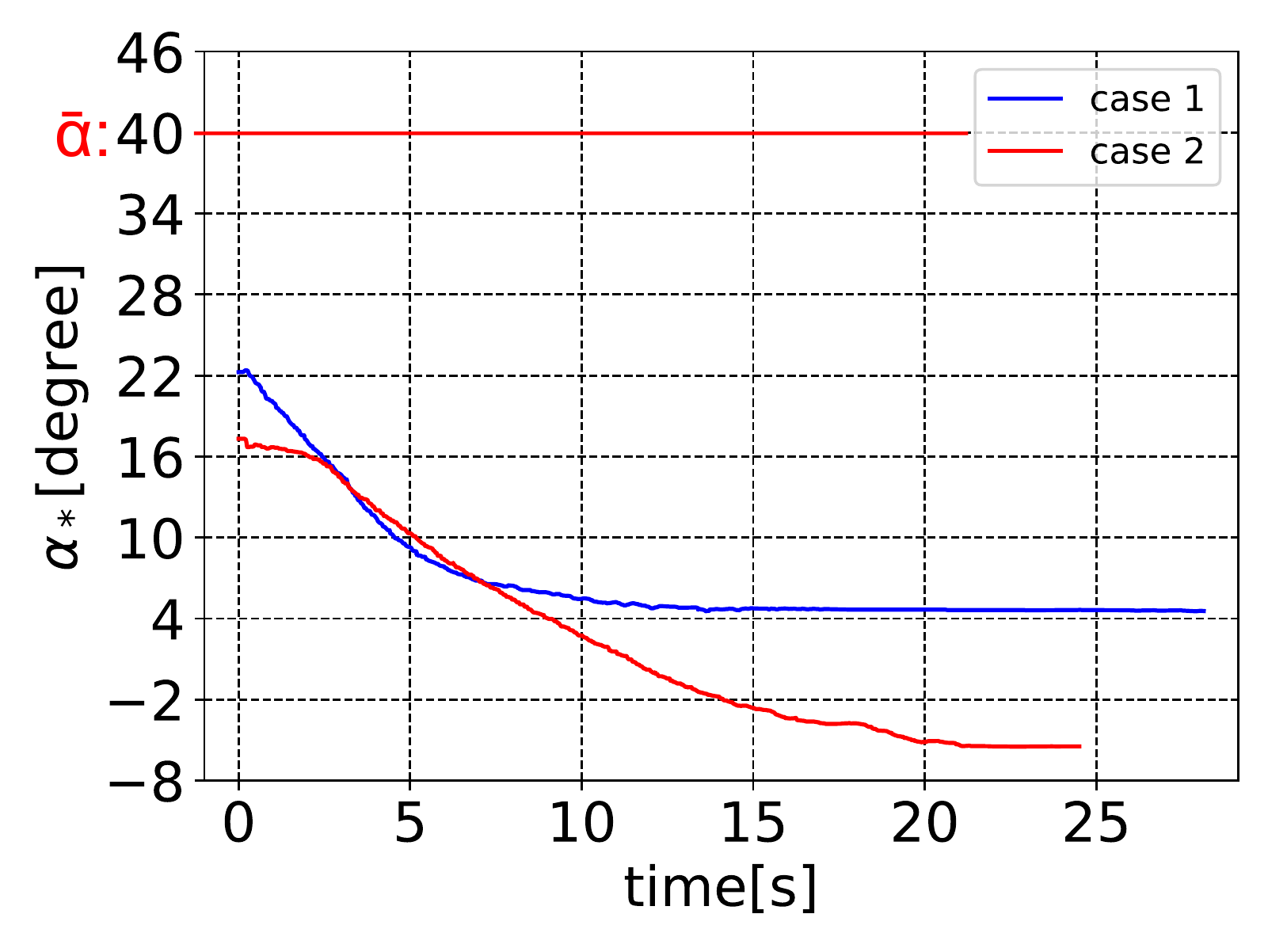}%
		\label{fig:AlphaStar}}
		\subfigure[Robot's linear velocity $v$]{\includegraphics[width=0.45\linewidth]{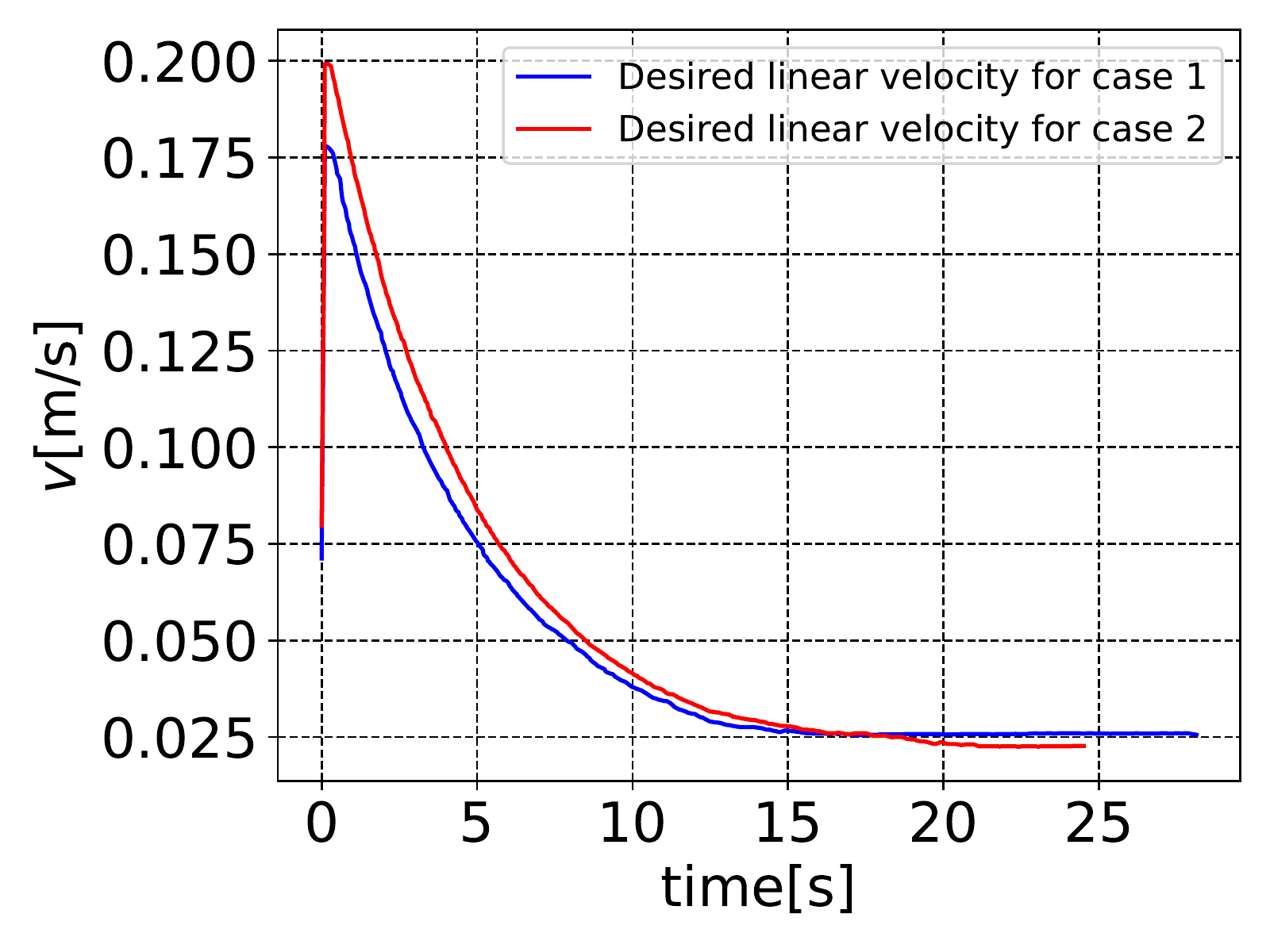}
		\label{fig:v}}
		\hfil
		\subfigure[Robot's angular velocity $w$]{\includegraphics[width=0.45\linewidth]{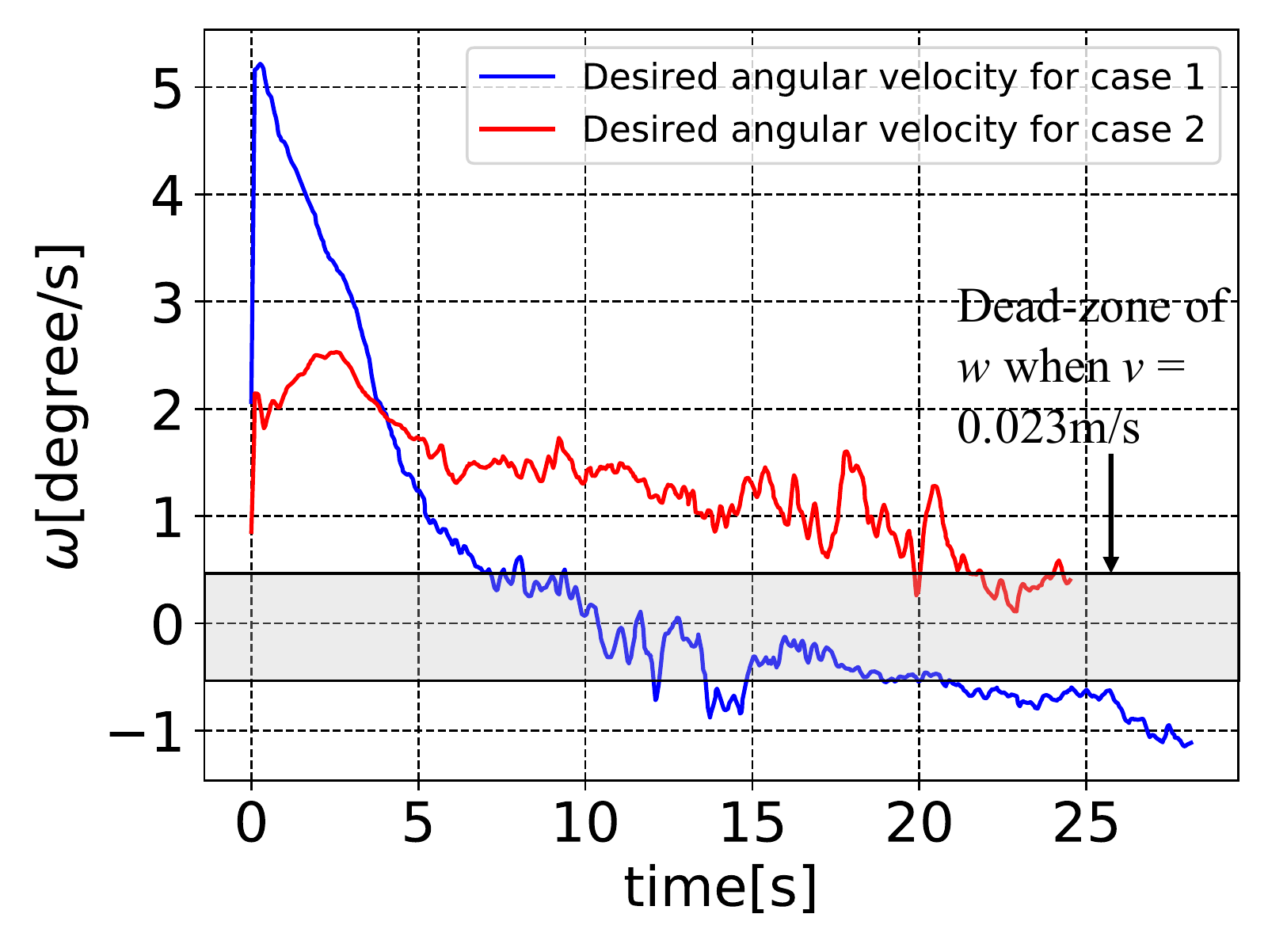}%
		\label{fig:w}}
      \caption{Example of experimental motion with the time progression of control variables, showing the convergence and constraints $\alpha^* < \bar{\alpha}$.} 
      \label{fig:T-V} 
   \end{figure}

\begin{figure}[t]
      \centering
  		\subfigure[case 1]{\includegraphics[width=0.47\linewidth]{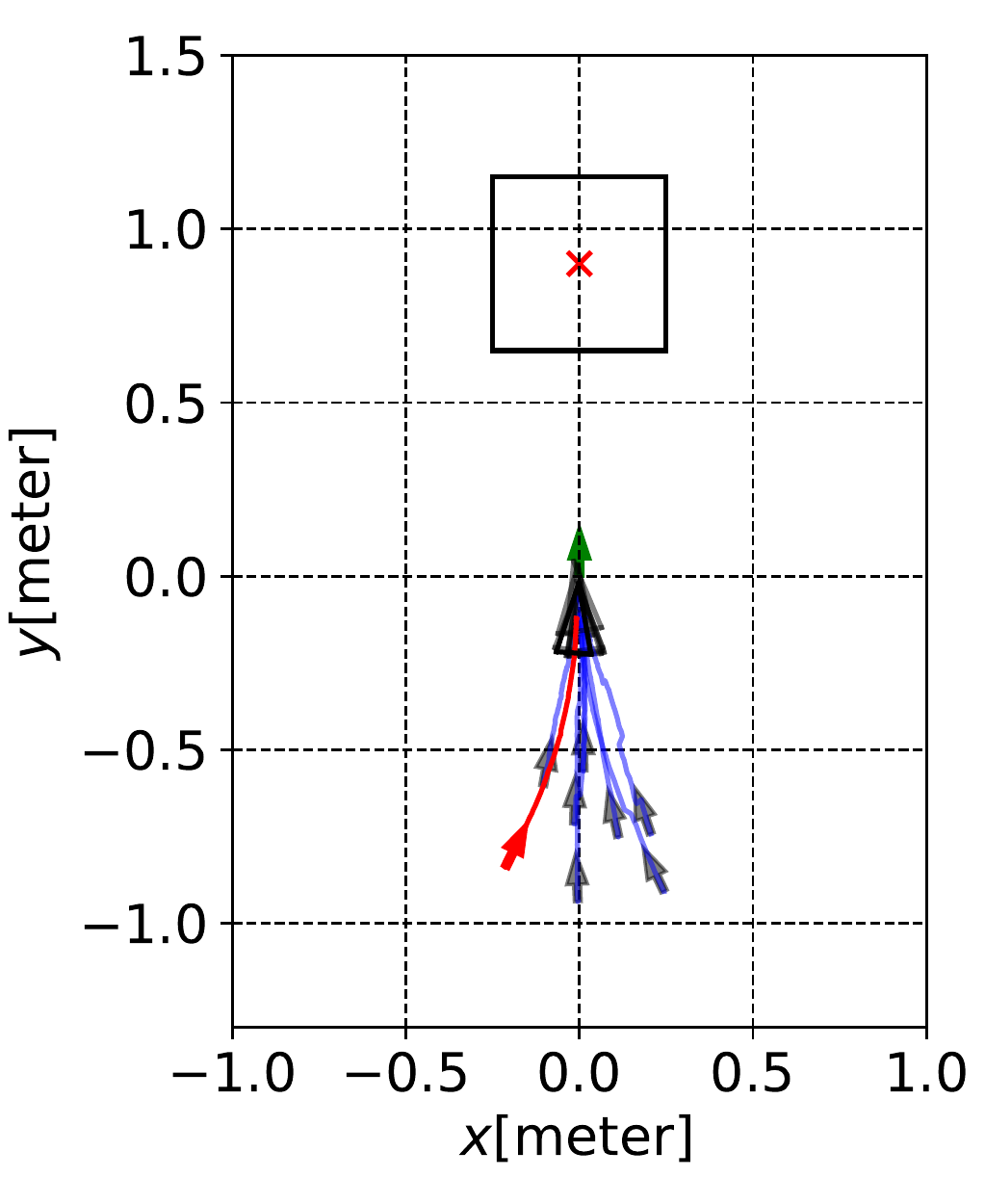}
		\label{fig:traj_}}
		\hfil
		\subfigure[case 2]{\includegraphics[width=0.47\linewidth]{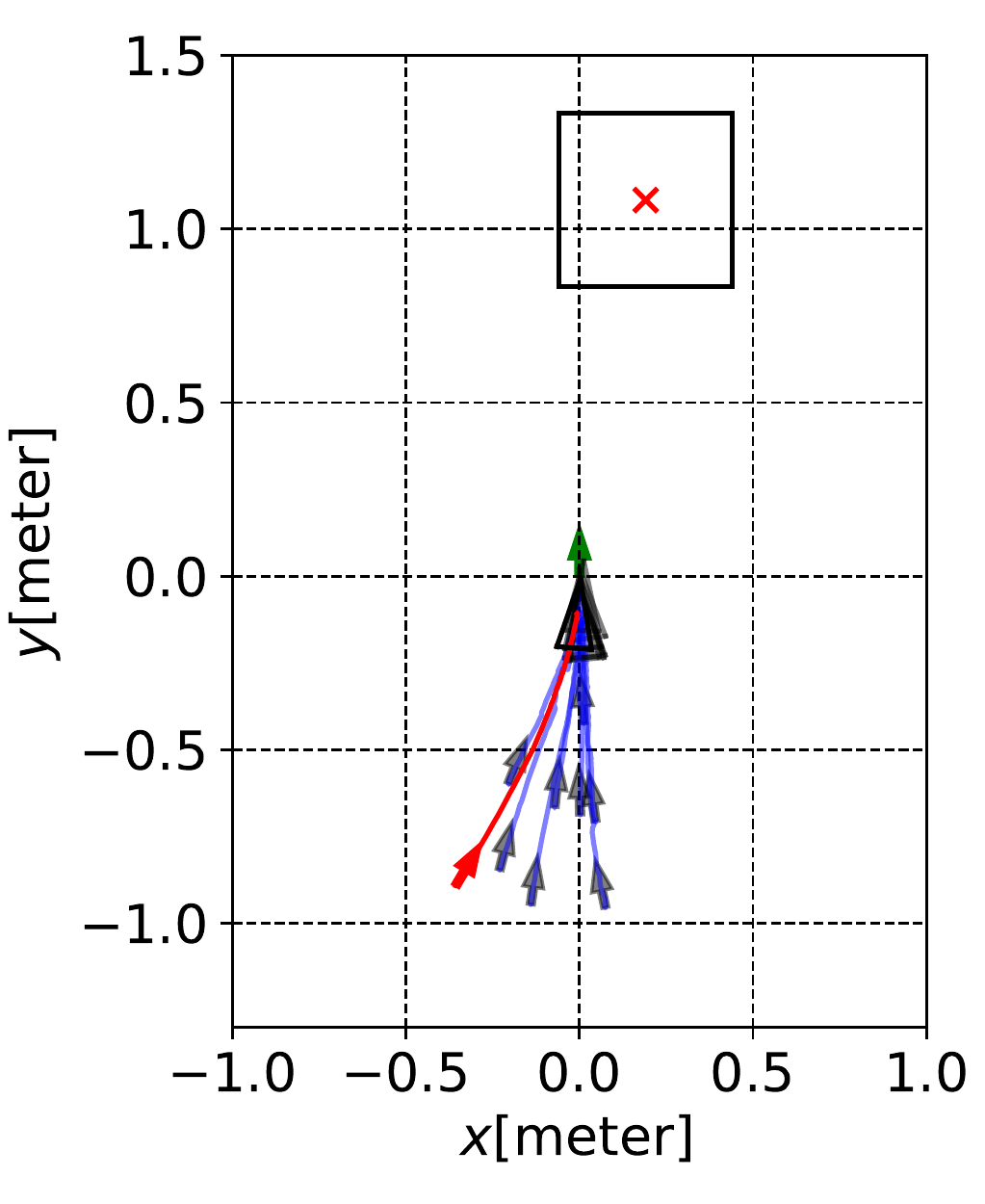}%
		\label{fig:traj_beta}}
      \caption{Robot's example trajectories. The sample cases in Fig. \ref{fig:T-V} correspond to initial states denoted with red arrows and the trajectories in red; Other tests with initial states denoted by black arrows, and trajectories in blue. Final states were denoted by black triangles.} 
      \label{fig:traj} 
   \end{figure}

In order to confirm the convergence of \((\rho, \alpha, \phi)\) and that \(\alpha^*\) never violates \(\bar{\alpha}\), one example for each case is shown in Fig. \ref{fig:T-V}, the trajectories of the robot are shown in Fig. \ref{fig:traj}. The plotted data were generated from the camera and control command of the system itself. For case 1, $\rho$, $\alpha$ and $\phi$ could converge to around 0.12m, -0.5 degrees and 4.5 degrees correspondingly. $\alpha^*$ decreased from 22.2 degrees to 4.6 degrees remaining within $\bar{\alpha}$. For case 2, $\rho$, $\alpha$ and $\phi$ could converge to around 0.11m, 1.8 degrees and 3.3 degrees correspondingly. $\alpha^*$ decreased from 17.3 degrees to -5.5 degrees remaining within $\bar{\alpha}$. 

\begin{figure}[t]
\begin{center}
\includegraphics[width=0.8\linewidth]{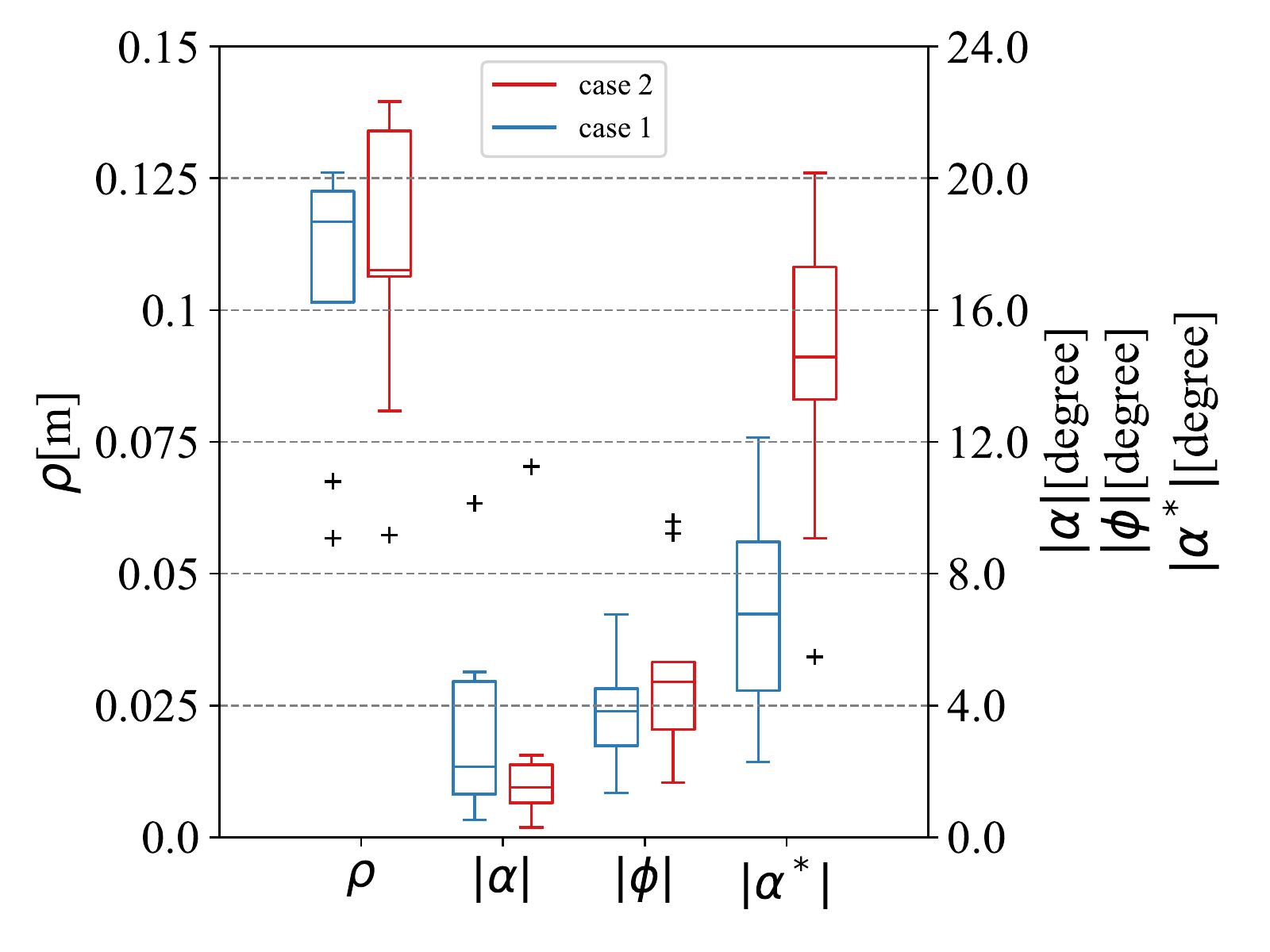}
\caption{Evaluation of convergence for multiple tests, showing the final states of the robot. The robot successfully  entered the safety region $\Omega$ of \((\rho < 0.15m, |\alpha| < 10^\circ, |\phi| < 10^\circ)\) in most cases.}
\label{multib}
\end{center}
\end{figure}

Overall results depicted in Fig. \ref{multib} describe the final states of the robot. For all cases $\rho$, $\alpha$ and $\phi$ entered the safety zone region defined above, the maximum $\alpha^*$ along the trajectory is $12.9^\circ \pm 6.7^\circ$ for case 1, and $16.1^\circ \pm 2.7^\circ$ for case 2, therefore we can confirm that $\alpha^*$ always remained within the $\bar{\alpha}$.

\section{Conclusion}\label{sec_con}
In this work, an autonomous docking support approach for assistive mobility devices was proposed. The problem of keeping a real object inside the FOV of a camera and docking to a virtual landmark that is inferred from the object itself was defined by a novel description with generalized virtual landmark and camera parameters. A simple but effective virtual landmark estimation algorithm is designed and verified by placing the robot in a number of different initial postures. We also proposed a nonlinear feedback controller considering the limited HFOV of the camera, the convergence of the system could be guaranteed under selected $k_1, k_2, k_3$ and given constrained feasible space.

The proposed method is general, computationally low-cost and fully applicable to any mobility device such as wheelchair and standing mobility device, and equally to any differential-driven mobile robot. An experiment indoor environment showed the robot docking to a chair successfully from multiple initial states. 
There are also a few limitations with the proposed control law regarding the feasible space, but it could be assisted by the user’s action, which would be considered as sharing control, in this case, an indicator of the feasible space might be needed for the user.

Another limitation of the system is the usage of the tracking camera, with an accuracy depending on the texture of the surrounding environment, but it could be easily replaced or fused with other sensor sources such as wheel encoders.
Future work will explore the proposed approach with user riding for evaluating the effectiveness of assistance during docking and the perception of the user to verify what needs could be mets.

\addtolength{\textheight}{-3cm}  


%



\bibliographystyle{IEEEtran}
\bibliography{IEEEabrv,bib_qolo_docking}
  \begin{IEEEbiography}[{\includegraphics[width=1in,height=1.25in,clip,keepaspectratio]{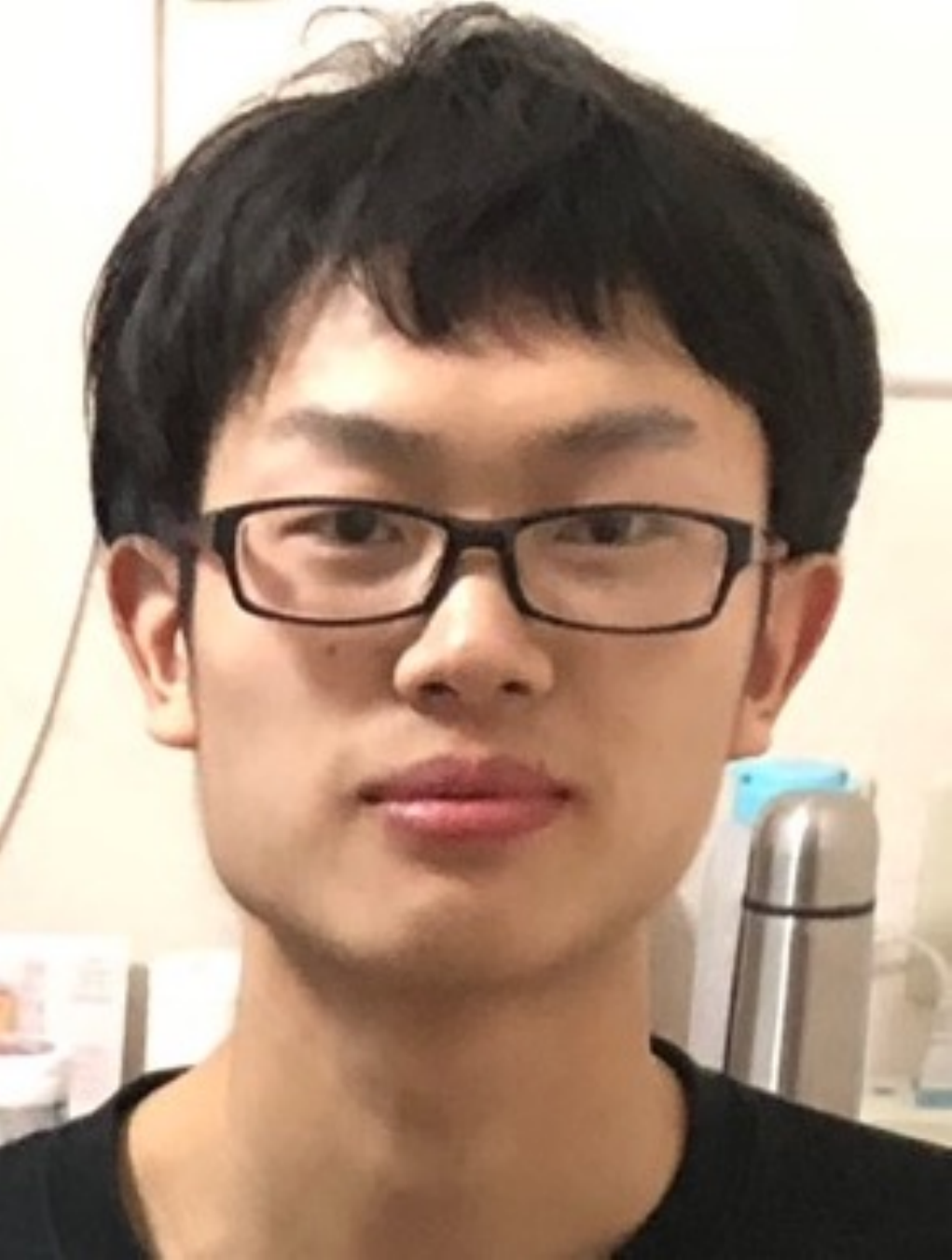}}]{Yang Chen}
received the B.Eng. in mechanical engineering from Jilin University, China, in 2017 and his master degree in human informatics  from the University of Tsukuba, Japan, in 2020 where he is currently working toward the Ph.D. degree. His research interests include assistive robotics, smart mobility and robot control interface. 
  \end{IEEEbiography}
  
\begin{IEEEbiography}[{\includegraphics[width=1in,height=1.25in,clip,keepaspectratio]{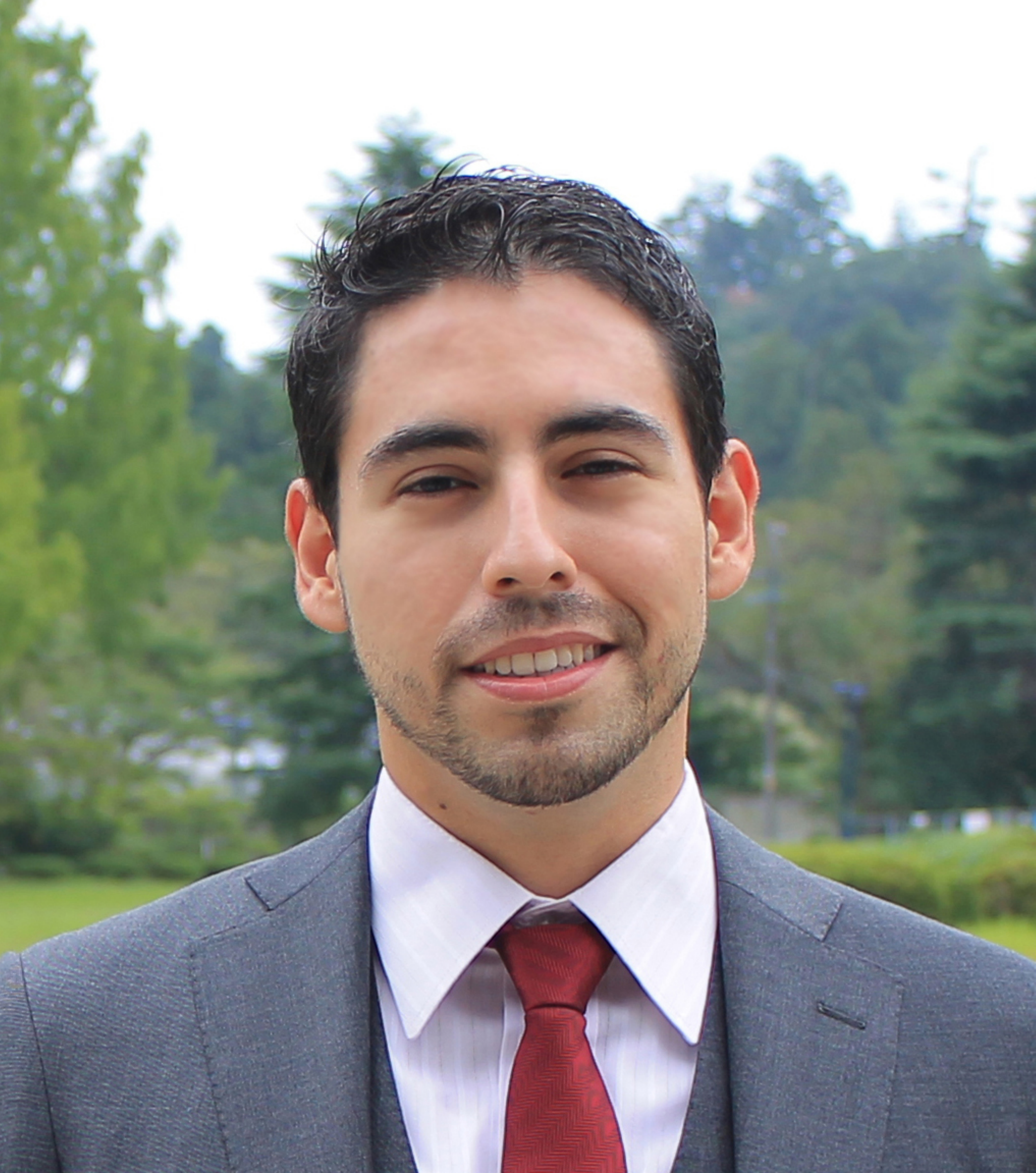}}]{Diego F. Paez-Granados}

received his Ph.D. in Bioengineering and Robotics from Tohoku University in 2017; for his work on physical guidance and skill teaching by a robot. Currently, he is a researcher at the Learning Algorithms and Systems Laboratory at EPFL in Switzerland, and visiting researcher at the University of Tsukuba, Japan. His research interests include physical and cognitive human modelling, compliance in human-robot interaction, shared-control for robot navigation, and soft-robot design and control with human-in-the-loop.

\end{IEEEbiography}

  \begin{IEEEbiography}[{\includegraphics[width=1in,height=1.25in,clip,keepaspectratio]{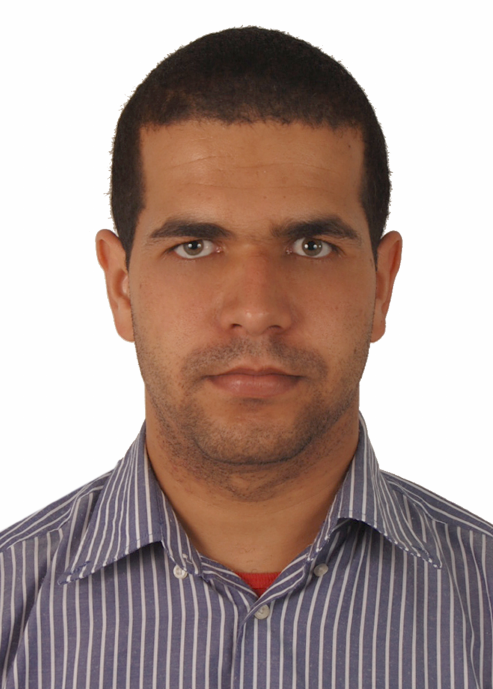}}]{Bruno Leme}
  received his Ph.D. degree in engineering from the Graduate School of Systems and Information Engineering, University of Tsukuba, Japan, in 2020. He is currently a researcher in the Artificial Intelligence Laboratory, University of Tsukuba, Japan. His research interests include assistive technologies for gait training, behavior estimation using social cues, and social robotics.
  \end{IEEEbiography}
  
\begin{IEEEbiography}[{\includegraphics[width=1in,height=1.25in,clip,keepaspectratio]{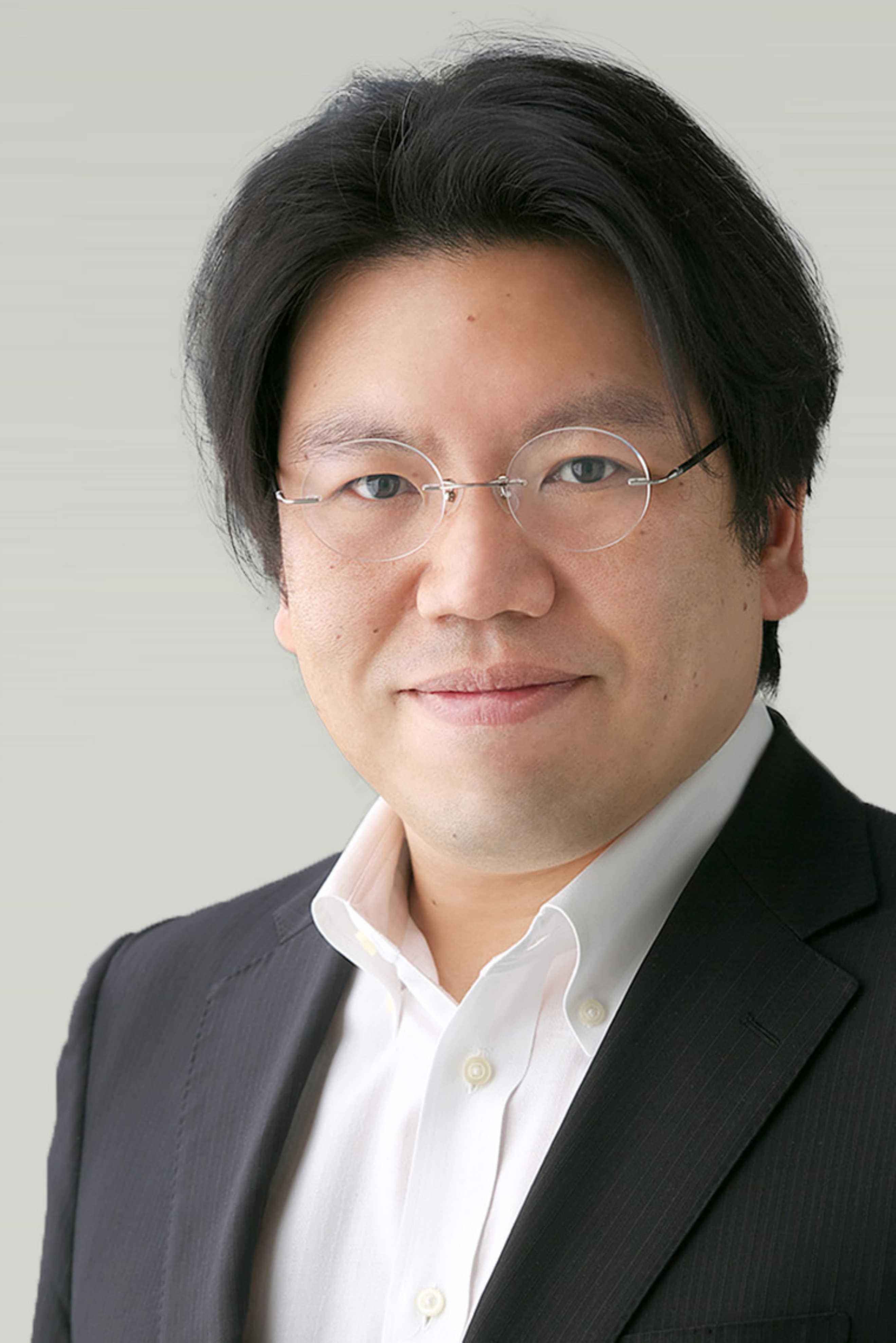}}]{Kenji Suzuki}
received the Ph.D. degree in pure and applied physics from Waseda University, Tokyo, Japan, in 2003. He is currently a full Professor with the Center for Cybernics Research and the principal investigator of Artificial Intelligence Laboratory, University of Tsukuba, Japan. His research interests include wearable robotics and devices, affective computing, social robotics, and assistive robotics. He is a member of IEEE and ACM.
\end{IEEEbiography}
\end{document}